\documentclass[letterpaper, 10 pt, conference]{ieeeconf}  % Comment this line out if you need a4paper

\IEEEoverridecommandlockouts                              % This command is only needed if 
\usepackage{graphics} % for pdf, bitmapped graphics files
\usepackage{epsfig} % for postscript graphics files
\usepackage{amsmath} % assumes amsmath package installed
\usepackage{amssymb}  % assumes amsmath package installed
\usepackage{booktabs}
\usepackage{subcaption}
\usepackage{stfloats}
\usepackage{bm}
\usepackage{multirow}
\usepackage{flushend}
\renewcommand{\baselinestretch}{0.96}

\usepackage{xcolor}
\usepackage[normalem]{ulem}

\usepackage [final]{changes}

\title{\LARGE \bf
Distance Field Driven Reinforcement Learning for \\Reactive and Robust Collision Avoidance
}

\author{% should be anonymous Taiga Ikeshima % <-this % stops a space
}

\author{
\added{Taiga Ikeshima$^{1}$,
Sheila Sutjipto$^{2}$,
Lan Wu$^{2}$,
Francisco J. Naranjo-Campos$^{3}$,}\\
\added{Taiki Ishigaki$^{1}$,
Taisuke Kobayashi$^{4}$,
Teresa A. Vidal-Calleja$^{2}$,
and Eiichi Yoshida$^{1}$}%
\thanks{\added{$^{1}$Taiga Ikeshima (8526501@ed.tus.ac.jp), Taiki Ishigaki, and Eiichi Yoshida (eiichi.yoshida@rs.tus.ac.jp) are with Tokyo University of Science, Japan.}}%
\thanks{\added{$^{2}$Sheila Sutjipto, Lan Wu, and Teresa A. Vidal-Calleja are with the University of Technology Sydney, Australia.}}%
\thanks{\added{$^{3}$Francisco J. Naranjo-Campos is with Universidad Carlos III de Madrid, Spain.}}%
\thanks{\added{$^{4}$Taisuke Kobayashi is with the National Institute of Informatics, Japan.}}%
}

\begin{document}

\maketitle
\thispagestyle{empty}
\pagestyle{empty}

%%%%%%%%%%%%%%%%%%%%%%%%%%%%%%%%%%%%%%%%%%%%%%%%%%%%%%%%%%%%%%%%%%%%%%%%%%%%%%%%
\begin{abstract}
%(updated)
%This study aims to control the motion of a robotic manipulator using Reinforcement Learning (RL). Specifically, by incorporating an environmental representation known as the Euclidean Distance Field (EDF) into RL, the goal is to utilize information recognized from the environment to reach the target position while avoiding obstacles. The EDF provides distance and gradient information from any given point to obstacles to be avoided. By integrating this into the observation space and reward terms in RL, the agent's behaviour can be determined. This enhances the agent's perception of the environment to improve the adaptability of obstacle avoidance.

%The effectiveness of the learning method was first verified by simulation using a redundant manipulator robot. Real-world experiments were then conducted using an RGB-D camera to calculate the EDF in real time, confirming the obstacle avoidance performance.

Robust and reactive collision avoidance remains a central challenge for deploying Reinforcement Learning (RL) policies on real robotic manipulators, particularly when transferring behaviours learned in simulation to the physical world. This paper presents an RL framework that integrates the Euclidean Distance Field (EDF) directly into the observation space, providing the agent with continuous geometric cues that support stable and responsive avoidance behaviours. By leveraging the rich spatial information encoded in the EDF, the method enables a streamlined reward formulation that achieves effective obstacle avoidance both with and without Artificial Potential Field guidance. The proposed approach is validated on a 7‑DoF redundant manipulator, demonstrating zero‑shot transfer from simulation to real hardware. Despite being trained exclusively with \added{randomised} static objects, the deployed policy generalizes to previously unseen real environments and exhibits robust online avoidance of dynamic obstacles. These results highlight the potential of EDF‑based representations to mitigate the sim‑to‑real gap and support reliable and real-time collision‑aware behaviour in complex and changing environments.

\end{abstract}
%%%%%%%%%%%%%%%%%%%%%%%%%%%%%%%%%%%%%%%%%%%%%%%%%%%%%%%%%%%%%%%%%%%%%%%%%%%%%%%%
\section{Introduction}
\label{sec:intro}

Generating motions that allow a robot to safely reach a target position while avoiding obstacles remains one of the fundamental challenges in mobile robotics and manipulation. As robots are increasingly deployed in cluttered, dynamic, and unstructured environments, the ability to reactively determine collision‑free actions from real‑time sensory data has become necessary for robust autonomy. 
This need for reliable, responsive obstacle avoidance is especially pressing where the robot must continuously resolve its motion when presented with incomplete spatial modelling.
%This transition shifts the challenge from path-following to dynamic reactivity, where the robot must continuously resolve its motion when presented with incomplete spatial modeling.

% Motion planning - usually offline, reactive or learning-based planning methods have been proposed
% In the real world - needs information about the environment - obstacle geometry and location - still an issue
% Recently, learning-based reactive motion has made progress. One is using APF with RL, but still getting the environment information is an issue.
% Distance field is a solution to obtain the environment information in real time and easy to implement in the learning scheme.
% In this research
% Contributions
% Integration of EDF, distance field with gradient into RL framework for robust and reactive collision avoidance
% Demonstrating its effectiveness through comparison with baseline methods
% Validating its advantages through the direct deployment in the real world

A classical solution to real‑time obstacle avoidance is the Artificial Potential Field (APF) method \cite{khatib1986potential}, which generates control actions by combining attractive and repulsive potentials. While APF is computationally efficient and reactive, it is known to suffer from local minima and non-reachable goals in the presence of obstacles, restricting its applicability in more complex environments. 
Beyond reactive-based strategies like APF, collision avoidance has also been addressed using offline motion planning by sampling-based methods such as Probabilistic Roadmaps (PRM)~\cite{Kavraki96} and Rapidly-exploring Random Trees (RRT) \cite{lavalle1998rapidly}. While these sampling-based approaches can navigate complex spaces, they often incur significant computational overhead and lack the real-time reactivity for operating in dynamic or unmodelled environments. 
% To address the need for real-time safety with the computational cost of global replanning, Control Barrier Functions (CBF)~\ref{} have emerged as a popular framework for ensuring real-time safety.
% Beyond APF, collision avoidance has also been addressed using offline motion planning by sampling-based methods such as the Rapidly-exploring Random Tree (RRT) \cite{lavalle1998rapidly}, which 
% samples random points in space, extends from the nearest node toward the sampled direction by a fixed distance, and 
% grows a search tree in the configuration space while checking collisions of the generated branches, as well as Probabilistic Roadmap (PRM) \cite{Kavraki96} which creates the network of collision-free connections between configrations in advance to search a path to the given start and goal. 
As an alternative method, optimisation-based motion planning \cite{zhang2020optimization} has been proposed, formulating collision avoidance as a constrained optimisation problem and directly computing a trajectory. 

% \begin{figure}[t]
%     \centering
%     \includegraphics[width=\columnwidth]{figures/PPO.png} % or width=\linewidth
%     \caption{Overview of the EDF-based RL pipeline.}
%     \label{fig:rl-pipeline}
% \end{figure}

% \begin{figure}
%     \centering
%     \includegraphics[width=\columnwidth]{edf-rl-overview.pdf}
%     \caption{Real-time collision avoidance using RL with Distance and Gradient Field}
%     \label{fig:placeholder}
% \end{figure}

% \begin{figure}
%     \centering
%     \includegraphics[width=\columnwidth]{edf-rl-overview -2.pdf}
%     \caption{Real-time collision avoidance using RL with Distance and Gradient Field}
%     \label{fig:placeholder}
% \end{figure}

% \begin{figure}
%     \centering
%     \includegraphics[width=1.0\linewidth]{Copia de Record Expert Place Trajectory-4.pdf}
%     \caption{Real-time collision avoidance using RL with Distance and Gradient Field}
%     \label{fig:placeholder}
% \end{figure}

\begin{figure}
    \centering
    \includegraphics[width=0.95\linewidth]{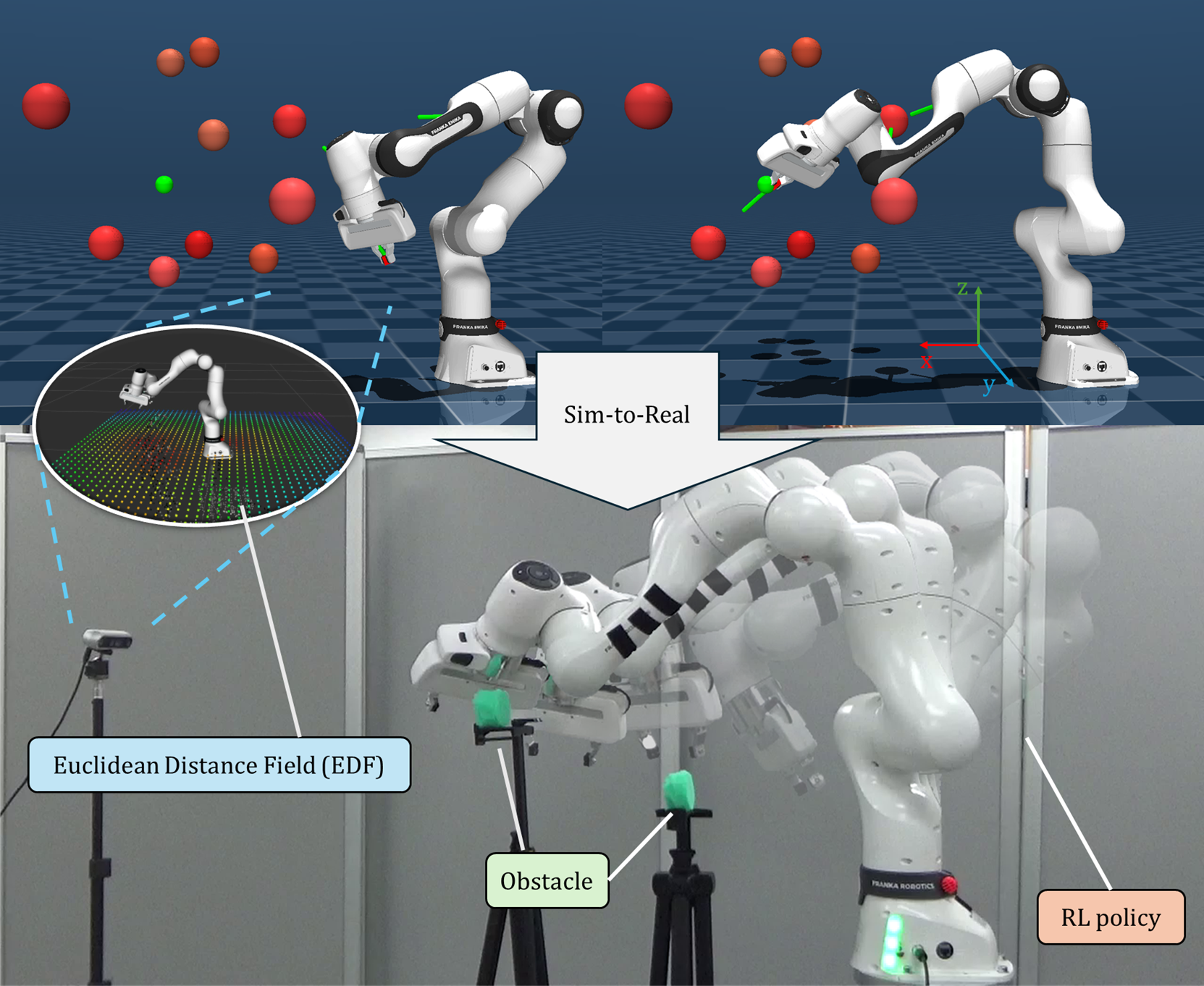}
    \caption{Real-time collision avoidance using RL with a Euclidean Distance and Gradient Field, learnt in simulation and transferred to real-robot experiments.}
    \label{fig:teaser}
\end{figure}

% Recent advances in machine learning have opened new directions in motion planning. Deep networks can encode the structure of the environment for collision‑free trajectory prediction \cite{Yip2020MPNet}, while Reinforcement Learning (RL) \cite{sutton1998reinforcement} enables robots to acquire reactive behaviours directly through interaction.
% Schneider et al. \cite{schneider2025robot} integrated APF into deep reinforcement learning to mitigate APF’s local‑minima issue; however, their method still assumes that accurate information about obstacle location and geometry is readily available.

Recent advances in machine learning have introduced alternative strategies for navigating complex environments. Deep networks can encode the structure of the environment for collision‑free trajectory prediction \cite{Yip2020MPNet}, while Reinforcement Learning (RL) \cite{sutton1998reinforcement} has enabled robots to acquire reactive behaviours directly through interaction with their simulated environment. Hybrid frameworks seek to leverage the benefits of combining traditional methods and RL. For instance, Schneider et al. ~\cite{schneider2025robot} integrated guidance from APFs into deep RL to mitigate the local‑minima issue inherent in classical potential fields. However, their method relies on the assumption that accurate spatial information regarding obstacle location and geometry is readily available. Xu \textit{et al.}~\cite{app14114936} proposed a hybrid architecture in which a deep RL policy produces avoidance actions that guide an APF planner online, enabling real‑time motion generation in cluttered scenes. They further extend the potential‑field formulation to account for link‑level proximity for full‑body collision avoidance. Li \textit{et al.}~\cite{article} similarly combine RL with APF by learning situation‑dependent scaling factors that modulate the magnitude of the field, improving robustness in cases where standard APF formulations struggle. These works highlight the appeal of APF‑RL combinations: APFs provide structure and computational efficiency, while RL introduces adaptivity and reduces the need for manual tuning.

% While those methods are useful for offline motion planning in complex environments,  
% % As a representative example of optimisation-based collision avoidance methods, 
% the Artificial Potential Field (APF) \cite{khatib1986potential}, which incorporates obstacles into the objective function within an unconstrained optimisation framework, still in . APF treats the robot as a point moving within a potential field, where the goal acts as an attractive source and obstacles act as repulsive sources. The robot is attracted toward the goal and repelled from obstacles, thereby moving toward the target position while avoiding collisions. Because APF has low computational cost, favourable real-time properties, and generates smooth obstacle-avoidance paths, it has been widely used as an obstacle avoidance method. However, when obstacles exist near the goal or when multiple obstacles are present, APF tends to fall into local optima, making efficient motion generation difficult. As a result, the robot may fail to reach the goal.

% In practice, obtaining precise and consistent environmental information is nontrivial and lies at the heart of the sim‑to‑real challenge. 
In practice, obtaining precise and consistent environmental information is nontrivial and remains a primary source of the sim-to-real gap for RL-based methods, as sensing and modelling can introduce noise and uncertainty that simulations fail to capture. 
Policies trained entirely in simulation may behave unpredictably when deployed if real‑world perception deviates from the training assumptions. So far, depth camera images and point clouds from LiDAR have shown a relatively small sim-to-real gap, but ad-hoc tricks are still needed to bridge the gap. Although domain randomisation \cite{8202133} helps by exposing the policy to varied conditions during training, reliable scene understanding remains essential for robust deployment.

%It is important to obtain information about the environment, such as the location and geometry of the obstacles, to apply either of the aforementioned methods for motion planning or learning-based motion behaviour. Indeed, the sim-to-real gap remains one of the main issues for those learning frameworks: even after training in simulations, the robot may encounter difficulty in deploying the learned policy as expected if there is a large difference from the real world. Previous research \cite{schneider2025robot} assumes that the environmental information can be acquired although it is not straightforward. While domain randomization \cite{xxx-domain-randomization} offers a workaround to address this issue by varying the parameters in the training phase, capturing the precise environmental information is essential. 

% domain randomization
% Advantage of EDF (Calionon for avoidance)  but not used in learning
% we get huge advantage by using EDF for RL 

To minimise these perceptual discrepancies, Euclidean Distance Fields (EDFs) have the potential to provide a robust representation for bridging the reality gap in reactive motion planning. 
Unlike raw point clouds, which are susceptible to sensor noise, or traditional geometric modelling of the environment, which often results in conservative behaviour, an EDF provides a dense spatial signal representing the distance to the nearest obstacle at any given point in the workspace. For example, Signed Distance Fields (SDFs) have been utilized for manipulator collision avoidance \cite{GOVONI2026105394, 8665083}. Despite its potential usefulness for learning, to the best of our knowledge, the distance field representations have not been well exploited with RL. 

An advantage of the distance field is its ability to be generated in real-time \cite{Ali2024IDMP}. By training RL policies on EDF-informed observations, an agent can learn to navigate based on the underlying spatial structure of the environment rather than specific pixel-level features or predefined location information. In this work, we propose to exploit EDFs with RL to reduce the sim-to-real gap, as the distance gradients provided by an EDF in the real world can be mapped consistently to the idealised fields encountered during training, ensuring more predictable and robust \added{behaviour} on real hardware. Fig.~\ref{fig:teaser} illustrates an example experiment for real-time collision avoidance using RL with EDF, trained in simulation and transferred to the real robot.
% In this context, the notion of distance field \cite{1634323} has a great potential to address the sim-to-real gap in reactive and robust motion planning and generation. A distance field can be defined in space to provide the distance and the gradient to the objects in the environment, whose typical instances are the signed distance field (SDF) \cite{curless1996volumetric} and Euclidean Distance Field (EDF) \cite{wu2021faithful}. The advantage of the distance field is that it can be obtained in real time using the recent progress of perception and data processing \cite{wu2021faithful}, leading to filling the sim-to-real gap.

The contributions of this paper are then as follows:
\begin{itemize}
    % \item A novel framework is proposed for RL to learn a policy for robust and reactive collision avoidance by integrating the EDF in the observation space.
    %\item A novel RL framework that integrates EDF into the observation space to facilitate robust and reactive collision avoidance.
    
    \item A novel RL framework that incorporates real-time Euclidean Distance Field measurements directly into the observation space, enabling the policy to learn reactive and robust collision‑avoidance behaviours %informed by real‑time geometric information.
    % \item The proposed framework ensures the performance of collision avoidance while simplifying the reward design compared to the previous work on APF-based RL that needed an additional term to avoid local minima
    \item A streamlined reward mechanism that achieves obstacle avoidance with and without APF guidance for the RL method. 
    % \item The effectiveness of the proposed EDF-based RL framework against the sim-to-real gap is validated through real-world experiments by deploying the learned policy to the redundant manipulator.

    \item Zero‑shot transfer from simulation to a 7-DoF redundant manipulator, where the deployed policy generalizes to real scenes and exhibits online avoidance of previously unseen \emph{dynamic} obstacles, despite being trained only with static objects in simulation.
    
    %\item Experimental validation of the framework's robustness against the sim-to-real gap, demonstrated through the deployment of the trained policy on a 7DoF redundant manipulator. 
\end{itemize}

\section{Preliminaries}\label{section:prelim}

\subsection{Reinforcement Learning}

RL~\cite{sutton1998reinforcement} provides a framework in which an agent learns a policy through interaction with an environment.
RL is commonly formulated as a Markov Decision Process (MDP) defined by $(\mathcal{S}, \mathcal{A}, P, R, \gamma)$, where $\mathcal{S}$ and $\mathcal{A}$ denote the state and action spaces, $P(\mathbf{s}_{t+1} \mid \mathbf{s}_t, \mathbf{a}_t)$ is the transition probability, $R(\mathbf{s}_t, \mathbf{a}_t)$ is the reward function, and $\gamma \in (0,1)$ is the discount factor.
However, although the environment has an underlying state $\mathbf{s}_t \in \mathcal{S}$ at each time step $t$, the agent cannot directly access it.
Instead, the agent receives an observation $\mathbf{o}_t \in \mathcal{O}$, from which it selects an action $\mathbf{a}_t \in \mathcal{A}$ according to a policy $\pi(\mathbf{a}_t\mid \mathbf{o}_t)$.
The environment then returns a scalar reward $r_t$ and transitions to a new state, producing the next observation $\mathbf{o}_{t+1}$.
Through this interaction loop, the agent improves its policy to maximise the expected cumulative reward.

% RL is commonly formulated as a Markov Decision Process (MDP) defined by $(\mathcal{S}, \mathcal{A}, P, R, \gamma)$, where $\mathcal{S}$ and $\mathcal{A}$ denote the state and action spaces, $P(s_{t+1} \mid s_t, a_t)$ is the transition probability, $R(s_t, a_t)$ is the reward function, and $\gamma \in (0,1)$ is the discount factor. 
% When the state is not fully observable, the problem is more precisely described as a Partially Observable Markov Decision Process (POMDP).

The return from time step $t$ is
\begin{equation}
G_t = \sum_{k=0}^{\infty} \gamma^k r_{t+k},
\end{equation}
and the objective is to find a policy that maximises $\mathbb{E}[G_t]$.

% If the observation space $\mathcal{O}$ encompasses the state space $\mathcal{S}$, RL can optimize the policy.
% Otherwise, the agent may fail to correctly predict the expected return and risk converging to wrong or locally optimal policies.
% To avoid this issue, it is desirable to design appropriate observations and avoid adding reward terms that depend on unobservable states.

\subsection{Proximal Policy Optimisation}

Let us adopt the Proximal Policy Optimisation (PPO)~\cite{schulman2017proximal} approach for agent training. PPO is a policy gradient method that stabilises learning by limiting excessive policy updates through a clipping mechanism.
% , avoiding the complex constrained optimisation required in TRPO~\cite{schulman2015trust}.
The policy $\pi_\theta(\mathbf{a}_t\mid\mathbf{o}_t)$ is updated using an advantage estimate $\hat{A}_t$, typically computed via Generalised Advantage Estimation (GAE), and an importance ratio $r_t(\theta)=\frac{\pi_\theta(\mathbf{a}_t \mid \mathbf{o}_t)}{\pi_{\theta_{\mathrm{old}}}(\mathbf{a}_t \mid \mathbf{o}_t)}$.
The clipped surrogate objective
\begin{equation}
\resizebox{0.90\columnwidth}{!}{$
\displaystyle
L^{\mathrm{PPO}}(\theta)=
\hat{\mathbb{E}}_t
\left[
\min\left(
r_t(\theta)\hat{A}_t,\,
\mathrm{clip}\left(r_t(\theta),1-\epsilon,1+\epsilon\right)\hat{A}_t
\right)
\right]
$}
\end{equation}
restricts large deviations from the previous policy, enabling stable and computationally efficient training.

\subsection{Euclidean Distance Field}\label{ssection:edf}
A robot's environment can be modeled as a surface or manifold $S\subset\mathbb{R}^D$ with boundary $\partial S$ whose orientation is given by surface normals. The Euclidean Distance Field (EDF) representation of $S$ is then a scalar field with value at $\mathbf{x}\in\mathbb{R}^D$ giving the distance to the closest point on the boundary $\mathbf{b}\in\partial S$,
\begin{equation}\label{eq:bg:edf}
    d(\mathbf{x}) = \min_{\mathbf{b}\in\partial S} \|\mathbf{x} - \mathbf{b}\|.
\end{equation}
 A key property of the EDF is that its gradient has unit norm for all $\mathbf{x} \in S \setminus \partial S$. That is, the gradient satisfies the \textit{Eikonal equation}, $\|\nabla d(\mathbf{x})\| = 1$.
The relevance of the Eikonal equation for distance field reconstruction is that it provides a strict condition for any reconstruction of the EDF to ensure the environmental representation. Note that any solution to the Eikonal equation is naturally a distance field. 
% The EDF~\cite{wu2021faithful} represents the workspace as a continuous scalar field that provides, for any query point in space, the Euclidean distance to the nearest obstacle surface and its spatial gradient.

% Let the obstacle region be a set $S \subset \mathbb{R}^D$ with boundary $\partial S$. The EDF distance function is defined as
% \begin{equation}
% d(\mathbf{x}) = \min_{y \in \partial S} \|\mathbf{x}-y\|.
% \end{equation}
% This distance can be interpreted as the arrival time of a unit-speed wavefront emanating from $\partial S$, and therefore satisfies the Eikonal equation
% \begin{equation}
% \|\nabla d(\mathbf{x})\| = 1.
% \end{equation}
% Geometrically, $\nabla d(\mathbf{x})$ points in the direction of steepest distance increase, which corresponds to the outward unit normal away from the obstacle surface. Hence, the gradient naturally provides an avoidance direction indicating how the robot can most efficiently move away from obstacles.

In this work, we adopt the Interactive Distance Field Mapping and Planning (IDMP) framework~\cite{Ali2024IDMP}, which provides an online, continuous EDF and gradient field representation based on Gaussian Processes (GPs)~\cite{wu2021faithful,le2023accurate}. IDMP incrementally constructs a fused distance field that supports real-time distance and gradient updates to allow navigation with dynamic obstacles.

Let $\mathcal{P}_i = \{\mathbf{x}_j\}_{j=1}^{N_i}$ denote the point cloud observed at time $i$. A temporary latent GP model, referred to as the \emph{Frustum Field}, is constructed using only the current measurements. The GP models a latent occupancy function
\begin{equation}
o(\mathbf{x}) \sim \mathcal{GP}(0, k(\mathbf{x},\mathbf{x}')),
\end{equation}
where $k(\cdot,\cdot)$ is a covariance kernel. Given training observations $\mathbf{y}=\mathbf{1}$, the posterior mean at a query point $\mathbf{x}^*$ is
\begin{equation}
\hat{o}(\mathbf{x}^*) = k(\mathbf{x}^*, X)
\left(K(X,X) + \sigma^2 I\right)^{-1}\mathbf{1},
\end{equation}
where $X$ is the set of training points and $K(X,X)$ is the covariance matrix.

The EDF is recovered using a kernel-specific reverting function $r(\cdot)$,
\begin{equation}
\label{eq:dist}
\hat{d}(\mathbf{x}^*) = r\big(\hat{o}(\mathbf{x}^*)\big).
\end{equation}

The gradient field can be obtained analytically by applying the linear derivative operator to the GP posterior,
\begin{equation}
\label{eq:grad}
\nabla \hat{d}(\mathbf{x}^*) \propto \nabla \hat{o}(\mathbf{x}^*)
= \nabla k(\mathbf{x}^*, X)
\left(K(X,X) + \sigma^2 I\right)^{-1}\mathbf{1}.
\end{equation}

IDMP maintains a global \emph{Fused Field} by incrementally updating a set of training points $\mathcal{P}_{0:i}$ across frames. Prior points are projected toward the inferred surface using the GP distance and gradient:
\begin{equation}
\hat{\mathbf{x}} = \mathbf{x} - \hat{d}(\mathbf{x})\,\nabla \hat{d}(\mathbf{x}),
\end{equation}
which enables efficient fusion and dynamic object removal without explicit data association.

This formulation produces a continuous and differentiable distance field that can be queried at arbitrary locations. As a result, IDMP naturally supports gradient-based motion planning and reactive control by directly providing both distance and gradient information in real time.

\begin{figure}[t]
    \centering
    \begin{subfigure}{.5\linewidth}
        \centering
        \includegraphics[width=.95\linewidth]{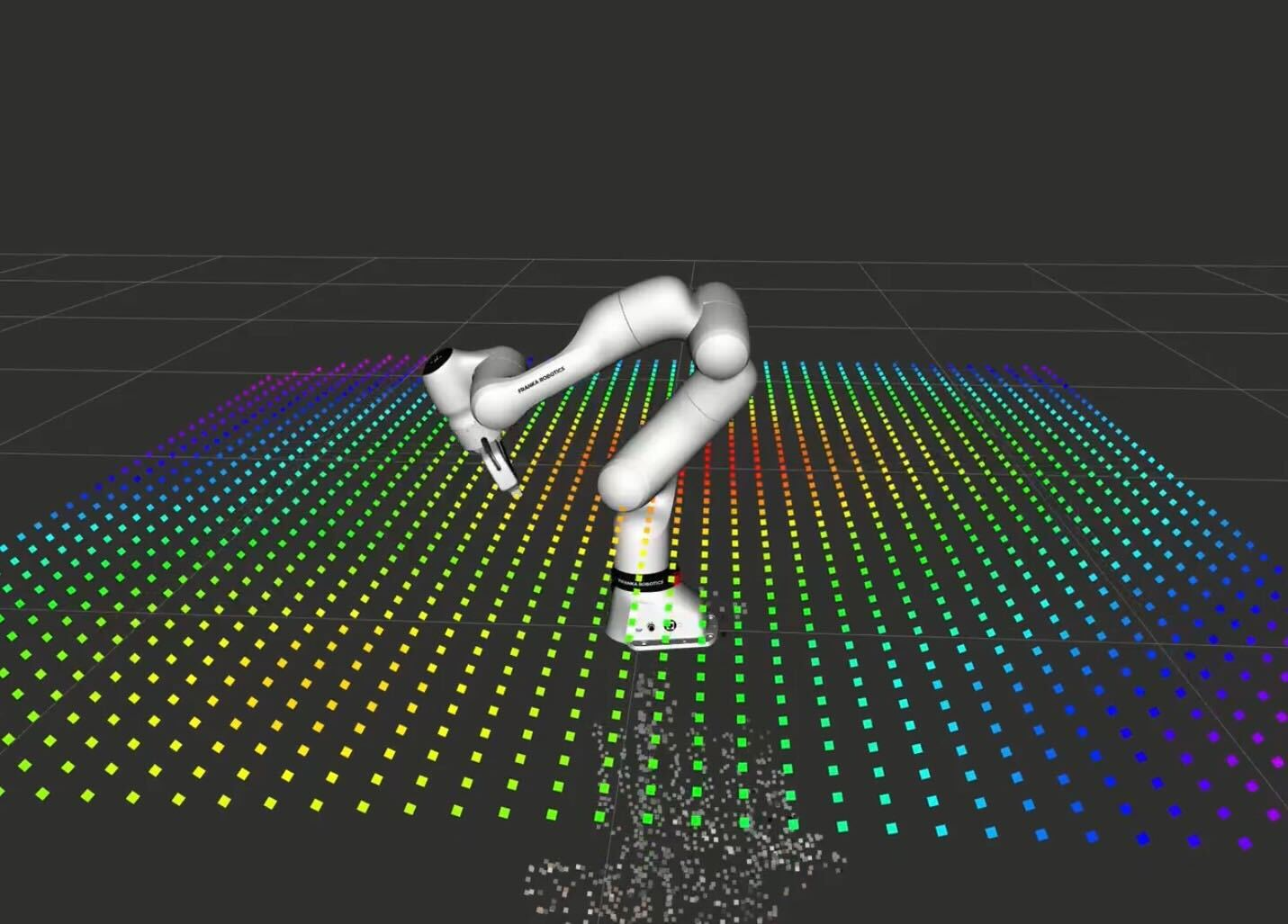}
        \caption{}
        \label{fig:edf_noobs}
    \end{subfigure}%
    \begin{subfigure}{.5\linewidth}
        \centering
        \includegraphics[width=.95\linewidth]{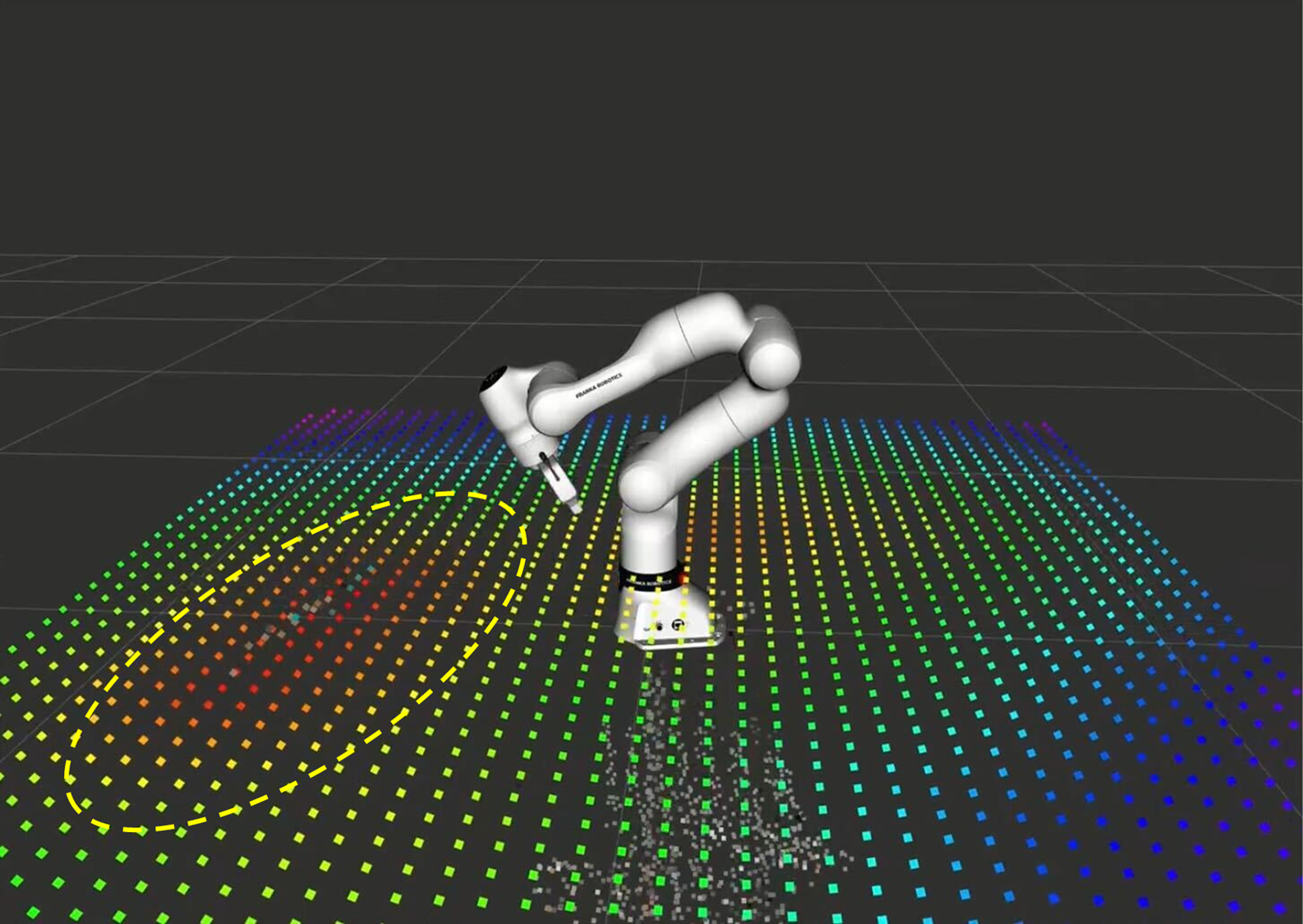}
        \caption{}
        \label{fig:edf_obs}
    \end{subfigure}%
    \caption{Cross-sectional $xy$-plane slice of the Euclidean Distance Field (EDF). (a) The EDF in an unobstructed environment. (b) The same environment with a single obstacle; the resulting change in the distance field is highlighted by the red points within the yellow dashed outline. }
    \label{fig:edf_example}
\end{figure}

Fig.~\ref{fig:edf_example} (a) and (b) show the spatial representation induced by the EDF.
The colour indicates the value of the distance field, where warmer colours correspond to smaller distances between the query location and the object.
In the obstacle-free scene, the distance distribution changes smoothly except in the vicinity of the robot arm.
In contrast, with an obstacle present, Fig.~\ref{fig:edf_example} (b) shows a locally intensified region (highlighted by the yellow dashed outline), indicating the obstacle and the reduced distance in its neighbourhood.

% \begin{figure}[t]
%     \centering
%     \begin{minipage}[t]{0.48\linewidth}
%         \centering
%         \includegraphics[width=\linewidth]{figures/EDFnoobs.png}
%         \caption{Environment w/o obstacles.}
%         \label{fig:edf_noobs}
%     \end{minipage}\hfill
%     \begin{minipage}[t]{0.48\linewidth}
%         \centering
%         \includegraphics[width=\linewidth]{figures/EDFobs.png}
%         \caption{Environment with obstacles.}
%         \label{fig:edf_obs}
%     \end{minipage}
% \end{figure}

\subsection{Artificial Potential Field (APF)}

The APF method~\cite{khatib1986potential} models robot motion as movement in a virtual force field.
Let $\mathbf{x}\in\mathbb{R}^3$ denote a point of interest on the robot's structure, such as the end effector (EE). The potential field is defined as
\begin{equation}
U(\mathbf{x})=U_{\mathrm{attr}}(\mathbf{x})+U_{\mathrm{rep}}(\mathbf{x}),
\end{equation}
composed of an attractive term toward the goal and a repulsive term from obstacles.

The attractive potential toward the goal $\mathbf{x}_g\in\mathbb{R}^3$ is defined as 
\begin{equation}
U_{\mathrm{attr}}(\mathbf{x})=\frac{1}{2}k_{\mathrm{attr}}\|\mathbf{x}-\mathbf{x}_g\|^2,
\end{equation}
where $k_{\mathrm{attr}}>0$.

The repulsive potential for obstacle avoidance is defined as
\begin{equation}
U_{\mathrm{rep}}(\mathbf{x}) =
\begin{cases}
\displaystyle
\frac{1}{2} k_{\mathrm{rep}}
\left(
\frac{1}{\rho(\mathbf{x})} - \frac{1}{\rho_0}
\right)^2,
& \rho(\mathbf{x}) \le \rho_0, \\
0,
& \rho(\mathbf{x}) > \rho_0,
\end{cases}
\end{equation}
where $k_{\mathrm{rep}} > 0$ is the repulsive gain, $\rho(\mathbf{x})$ is the distance to the nearest obstacle, and $\rho_0$ defines the influence range.
The repulsive potential increases rapidly as the robot approaches an obstacle and vanishes outside the influence range.

The artificial force acting on the robot is given by the negative gradient of the potential:
\begin{equation}
\mathbf{f}(\mathbf{x}) = -\nabla U(\mathbf{x})
= \mathbf{f}_{\mathrm{attr}}(\mathbf{x}) + \mathbf{f}_{\mathrm{rep}}(\mathbf{x}),
\end{equation}
where $\mathbf{f}_{\mathrm{attr}}(\mathbf{x}) = -\nabla U_{\mathrm{attr}}(\mathbf{x})$ and
$\mathbf{f}_{\mathrm{rep}}(\mathbf{x}) = -\nabla U_{\mathrm{rep}}(\mathbf{x})$.
This force determines the motion direction, enabling the robot to reach the goal while avoiding obstacles.

Additionally, to mitigate local minima and facilitate smooth navigation around obstacles, we incorporate the orthogonal compensation component $\mathbf{f}_{\perp}$~\cite{schneider2025robot} with a coefficient $\gamma \ll 1$.

\begin{equation}
\mathbf{f}_{\perp}=
\frac{\mathbf{f}_{\mathrm{attr}}\times \mathbf{f}_{\mathrm{rep}}}
{\lVert \mathbf{f}_{\mathrm{attr}}\times \mathbf{f}_{\mathrm{rep}}\rVert},
\qquad
\mathbf{f}_{\mathrm{adj}}=
\mathbf{f}_{\mathrm{attr}}+
\mathbf{f}_{\mathrm{rep}}+
\gamma \mathbf{f}_{\perp}.
\end{equation}

\section{Proposed Method}
\label{chap:method1}
%This chapter describes the configuration of a method that incorporates the Euclidean Distance Field (EDF) into deep reinforcement learning as an environmental representation.
This section presents the proposed control framework that integrates the EDF as the primary environmental representation within a Deep RL (DRL) architecture.

% \subsection{Reinforcement Learning Method Using EDF}

Let us represent the surrounding workspace using the IDMP framework~\cite{Ali2024IDMP}. As established in Section~\ref{section:prelim}, IDMP constructs \added{a continuous} scalar field $d(\mathbf{x})$ over the workspace such that $d(\mathbf{x})$ encodes the Euclidean distance from an arbitrary query point $\mathbf{x}$ to the nearest obstacle surface. Both $d(\mathbf{x})$ and its gradient $\nabla d(\mathbf{x})$ can be efficiently queried; the former quantifies proximity, while the latter offers a principled escape vector aligned with the outward surface normal of the nearest obstacle.

% We incorporate the EDF-derived distance and gradient information into both the reward design and the observation space of the RL framework.
\added{We incorporate EDF-derived distance and gradient information into the observation space and into the reward design for the APF-guided configurations.} By explicitly providing continuous proximity measures and avoidance-direction cues, the agent can select actions while directly accounting for obstacle distance and escape direction, enabling obstacle-aware policy learning in a physically meaningful and differentiable manner.

\subsection{Pipeline of the RL Method Using EDF}

\begin{figure}[t]
    \centering
    \includegraphics[width=0.95\columnwidth]{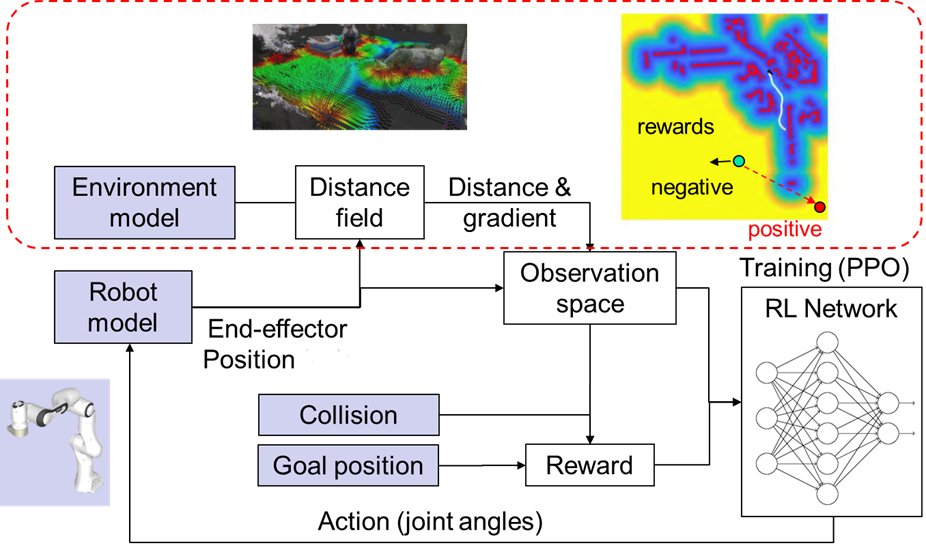} % or width=\linewidth
    \caption{Overview of the EDF-based RL pipeline.}
    \label{fig:rl-pipeline}
\end{figure}

% Figure~\ref{fig:rl-pipeline} shows the RL pipeline used in this study. Based on the end-effector position, the distance to obstacles in the environment and its gradient are computed, and these are provided as part of the observation space. The observation information and reward function are incorporated into a PPO-based RL network, and the robot outputs actions according to the resulting policy.

Fig.~\ref{fig:rl-pipeline} shows the proposed RL-based pipeline. The environment's EDF is generated from real-time observations using an RGB-D camera in the real world whereas it can be directly computed from the geometric information in the simulations. Query points for the distance and gradient to the nearest obstacle at the end of the robot's EE (and body links) are provided as part of the observation space in an RL framework. These observations are the input to a PPO-based policy network. For the APF-guided configurations, the reward function additionally incorporates APF guidance. The learned policy then generates actions that are executed by the robot with no requirement for fine-tuning in the real-world setup.
% , which is trained using a reward function that considers guidance from an APF. 

\subsection{PPO Implementation Details}
\label{subsec:ppo_details}
\added{
We trained all policies using PPO implemented in Stable-Baselines3 with a MultiInputPolicy. The actor and critic networks each consisted of two fully connected hidden layers with 64 units and Tanh activations. Training used a learning rate of $3.0 \times 10^{-4}$, a rollout length of 1024 steps, a mini-batch size of 64, 4 optimisation epochs per update, a discount factor of $\gamma=0.999$, a GAE parameter of $\lambda=0.98$, and an entropy coefficient of $0.01$.
}

% \subsection{Proximal Policy Optimisation}

% Let us adopt the Proximal Policy Optimisation (PPO)~\cite{schulman2017proximal} approach for agent training. PPO is a policy gradient method that stabilises learning by limiting excessive policy updates through a clipping mechanism, avoiding the complex constrained optimisation required in TRPO~\cite{schulman2015trust}.
% The policy $\pi_\theta(a \mid s)$ is updated using an advantage estimate $\hat{A}_t$, typically computed via Generalised Advantage Estimation (GAE), and an importance ratio $r_t(\theta)=\frac{\pi_\theta(a_t \mid s_t)}{\pi_{\theta_{\mathrm{old}}}(a_t \mid s_t)}$.
% The clipped surrogate objective
% \begin{equation}
% L^{\mathrm{PPO}}(\theta)=
% \mathbb{E}_t\!\left[
% \min\!\left(
% r_t(\theta)\hat{A}_t,\,
% \mathrm{clip}(r_t(\theta),1-\epsilon,1+\epsilon)\hat{A}_t
% \right)
% \right]
% \end{equation}
% restricts large deviations from the previous policy, enabling stable and computationally efficient training.

\subsection{Action Space}

Let us consider an action space vector that specifies incremental joint angle commands for the 7-DoF arm of the Franka Research 3 (FR3). At each time step, the agent outputs
\begin{equation}
\mathbf{a}_t = [\Delta q_1,\ldots,\Delta q_7]^\top \in \mathbb{R}^7,
\end{equation}
where $\Delta q_j$ denotes a small change applied to the $j$-th joint, for $j \in \{1, \dots, 7\}$. The subsequent joint target is computed as $\mathbf{q}_{t+1} = \mathbf{q}_t + \mathbf{a}_t$
% \begin{equation}
% \mathbf{q}_{t+1} = \mathbf{q}_t + \mathbf{a}_t,
% \end{equation}
with each joint angle clipped to its predefined range of motion to ensure physically feasible configurations. Since grasping is not considered, the gripper remains fixed and only the joints of the arm are controlled.

\subsection{Observation Space}
%Hamilton et al.\ report that the selection of observation features can substantially affect both learning efficiency and final policy performance~\cite{hamilton2025investigating}. 
%For our obstacle-avoidance task with randomly varying obstacle configurations, collision signals or penalties alone provide weak directional guidance. 
To supply informative and physically grounded avoidance cues, we include EDF-based features directly in the observation: the distance to the nearest obstacle, $d \in \mathbb{R}$, and the gradient of the distance field, $\nabla d  \in \mathbb{R}^3$, which are provided by querying the IDMP framework using Equations~\eqref{eq:dist} and~\eqref{eq:grad}, respectively.

The observation consists of the robot motion state, \added{the desired goal position,} and EDF-derived environmental information. We consider two conditions that differ in the spatial range where EDF features are queried.

% \added{In addition to these EDF-based observations, the desired goal position $\mathbf{p}_g \in \mathbb{R}^3$ is provided to the policy as an input.}

\subsubsection{EE-only}
EDF information is provided only at the robot EE. The observation is defined as
\begin{equation}
\mathbf{o}_t^{\mathrm{EE}}
=
\left[
\mathbf{p}_{ee}^\top,\ 
\dot{\mathbf{p}}_{ee}^\top,\ 
(\nabla d_{ee})^\top,\ 
d_{ee},\
\mathbf{p}_g^\top
\right]^\top
\in\mathbb{R}^{13},
\label{eq:obs-space-ee}
\end{equation}
where $\mathbf{p}_{ee}\in\mathbb{R}^3$ and $\dot{\mathbf{p}}_{ee}\in\mathbb{R}^3$ are the EE position and linear velocity, respectively. The scalar $d_{ee}\in\mathbb{R}$ is the EDF distance to the nearest obstacle evaluated at $\mathbf{p}_{ee}$, and $\nabla d_{ee}\in\mathbb{R}^3$ is the corresponding EDF gradient (pointing toward increasing clearance). \added{The desired goal position is denoted by $\mathbf{p}_g\in\mathbb{R}^3$.}

\subsubsection{Full-Body}
EDF information is additionally provided at a set of representative body points $\mathcal{P}=\{\mathbf{p}_i\}_{i=1}^{N}$. For each point $\mathbf{p}_i$, we compute the distance $d_i$ and gradient $\nabla d_i$ and append them to the observation:
\begin{equation}
\mathbf{o}_t^{\mathrm{FB}}
=
[(\mathbf{o}_t^{\mathrm{EE}})^\top,
\{(\nabla d_i)^\top\}_{i=1}^{N},
\{d_i\}_{i=1}^{N}]^\top
\in\mathbb{R}^{13+4N}.
\label{eq:obs-space-fb}
\end{equation}
This full-body (FB) formulation provides local clearance and escape-direction cues not only at the EE but also around intermediate links, enabling obstacle avoidance to be learned in a whole-body manner.

\subsection{Reward Function Design}

The reward jointly promotes goal reaching and obstacle avoidance, following Schneider et al.~\cite{schneider2025robot}:
% \begin{equation}
% R(s,a)=
% \begin{cases}
% -\lambda_c\,d_{\mathrm{ge}}+\mu(\cos\sigma-1), & \text{collision},\\
% -\lambda_{nc}\,d_{\mathrm{ge}}+\mu(\cos\sigma-1), & \text{otherwise}.
% \end{cases}
% \end{equation}
\begin{equation}
R(\mathbf{s},\mathbf{a})= -\lambda(\mathbf{s})\,d_{\mathrm{ge}}+\mu(\cos\sigma-1),
\label{eq:reward_function}
\end{equation}
\begin{equation}
\lambda(\mathbf{s})=
\begin{cases}
\lambda_c, & \text{collision},\\
\lambda_{nc}, & \text{otherwise}.
\end{cases}
\end{equation}
Here, $d_{\mathrm{ge}}$ is a penalising term based on the distance to the goal and $\cos\sigma$ encourages motion that is consistent with an APF reference direction. For configurations without APF guidance, the second term is omitted, and the reward reduces to $R(\mathbf{s},\mathbf{a})=-\lambda(\mathbf{s})d_{\mathrm{ge}}$. All hyperparameters are summarised in Table~\ref{tab:reward_params}. The reward definition is identical across observation settings.

\begin{table}[t]
\centering
\caption{Reward and APF guidance hyperparameters}
\label{tab:reward_params}
\renewcommand{\arraystretch}{1.15}
\setlength{\tabcolsep}{10pt}
\begin{tabular}{llc}
\toprule
Symbol & Meaning & Value \\
\midrule
$\mu$ & Direction-alignment term weight & $1.01$ \\
$\lambda_c$ & Goal-distance penalty (collision) & $2.0$ \\
$\lambda_{nc}$ & Goal-distance penalty (no collision) & $1.0$ \\
$w$ & Weight on repulsive term in APF & $0.2$ \\
$\gamma$ & Orthogonal escape term weight & $0.2$ \\
\bottomrule
\end{tabular}
\end{table}

\subsubsection{Goal-distance term}
The term $-\lambda(\mathbf{s}) d_{\text{ge}}$ penalises the agent's distance from the goal. 
% Now we will explain the goal-distance term $-\lambda(s) d_{\text{ge}}$ in (\ref{eq:reward_function}).
Let $\mathbf{p}_{ee}\in\mathbb{R}^3$ denote the EE position and $\mathbf{p}_g\in\mathbb{R}^3$ the goal position. The distance is defined as
\begin{equation}
d_{\mathrm{ge}} = \|\mathbf{p}_{ee} - \mathbf{p}_g\|.
\end{equation}
When a collision occurs, the \added{coefficient} of $d_{\mathrm{ge}}$ switches from $\lambda_{nc}$ to $\lambda_c$, penalising the agent for remaining in a collided state.

\subsubsection{Movement-direction consistency term}
The term $\mu(\cos\sigma-1)$ encourages the agent to align \added{its} motion with a collision-aware reference direction. 
% Now we will explain the movement-direction consistency term $\mu(\cos\sigma-1)$ in (\ref{eq:reward_function}).
We define the EE displacement between successive time steps, $t$ and $t-1$, as
\begin{equation}
\mathbf{v}_{ee}=\mathbf{p}_{ee, t}-\mathbf{p}_{ee, t-1},
\end{equation}
and construct an APF-inspired reference vector

\begin{equation}
% \mathbf{a}_{\mathrm{pf}}
\mathbf{v}_{\mathrm{ref}}
= w\,\frac{1}{d}\nabla d
+ (1-w)(\mathbf{p}_{g}-\mathbf{p}_{ee})
+ \gamma \hat{\mathbf{o}},
\end{equation}

where $d$ is the EDF evaluated at the query point providing the distance between the EE and nearest obstacle and $\nabla d$ is the corresponding gradient. The orthogonal escape term $\hat{\mathbf{o}}$ is
\begin{equation}
\hat{\mathbf{o}}
=
\frac{\mathbf{v}_{ee}\times\nabla d}{\|\mathbf{v}_{ee}\times\nabla d\|}.
\end{equation}

Finally,
\begin{equation}
\cos\sigma
=
% \frac{\mathbf{v}_{ee}\cdot\mathbf{a}_{\mathrm{pf}}}{\|\mathbf{v}_{ee}\|\,\|\mathbf{a}_{\mathrm{pf}}\|},
\frac{\mathbf{v}_{ee}\cdot\mathbf{v}_{\mathrm{ref}}}{\|\mathbf{v}_{ee}\|\,\|\mathbf{v}_{\mathrm{ref}}\|},
\end{equation}
and the term $\mu(\cos\sigma-1)\le 0$ is maximised when $\mathbf{v}_{ee}$ aligns with $\mathbf{v}_{\mathrm{ref}}$. Note that $d$ and $\nabla d$ are inferred using Equations~\eqref{eq:dist} and~\eqref{eq:grad}.

\begin{figure}[b]
    \centering
    \includegraphics[width=0.5\linewidth]{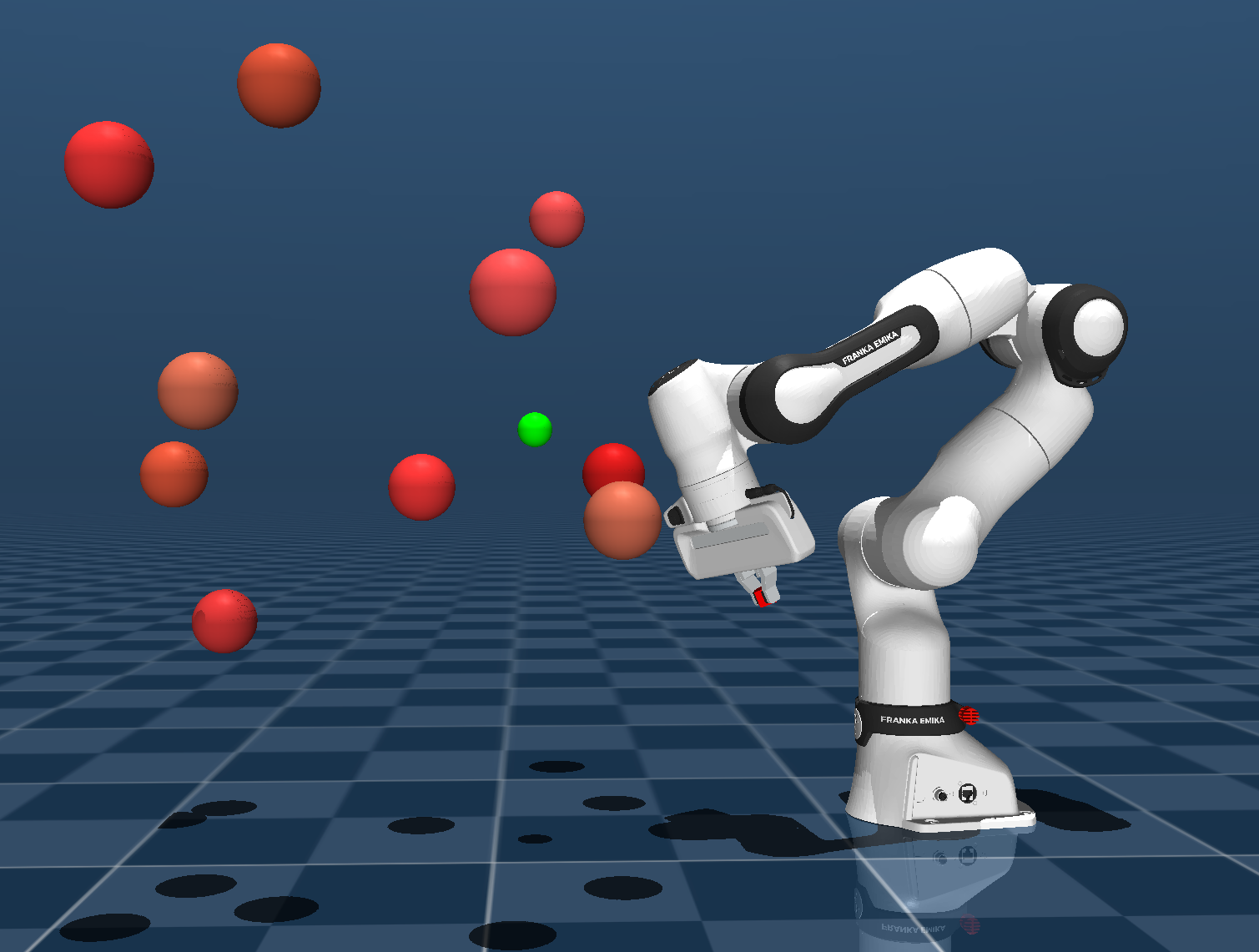}
    \caption{Simulation setup.}
    \label{fig:simulation_task}
\end{figure}

\begin{table}[b]
\centering
\caption{Sampling ranges used for obstacle and goal generation in simulation}
\label{tab:sampling_ranges_seeds}
\renewcommand{\arraystretch}{1.15}
\setlength{\tabcolsep}{10pt}
\begin{tabular}{lll}
\toprule
Category & Parameter & Range [min, max] \\
\midrule
                      & $x$ [m] & $[0.45,\ 1.00]$ \\
Obstacles             & $y$ [m] & $[-0.50,\ 0.50]$ \\
                      & $z$ [m] & $[0.20,\ 0.80]$ \\
% \addlinespace
                      & $x$ [m] & $[0.50,\ 0.80]$ \\
Goal                  & $y$ [m] & $[-0.30,\ 0.30]$ \\
                      & $z$ [m] & $[0.30,\ 0.50]$ \\
\bottomrule
\end{tabular}
\end{table}

\begin{figure}[b]
    \centering
    \begin{subfigure}{.5\linewidth}
        \centering
        \includegraphics[width=.85\linewidth]{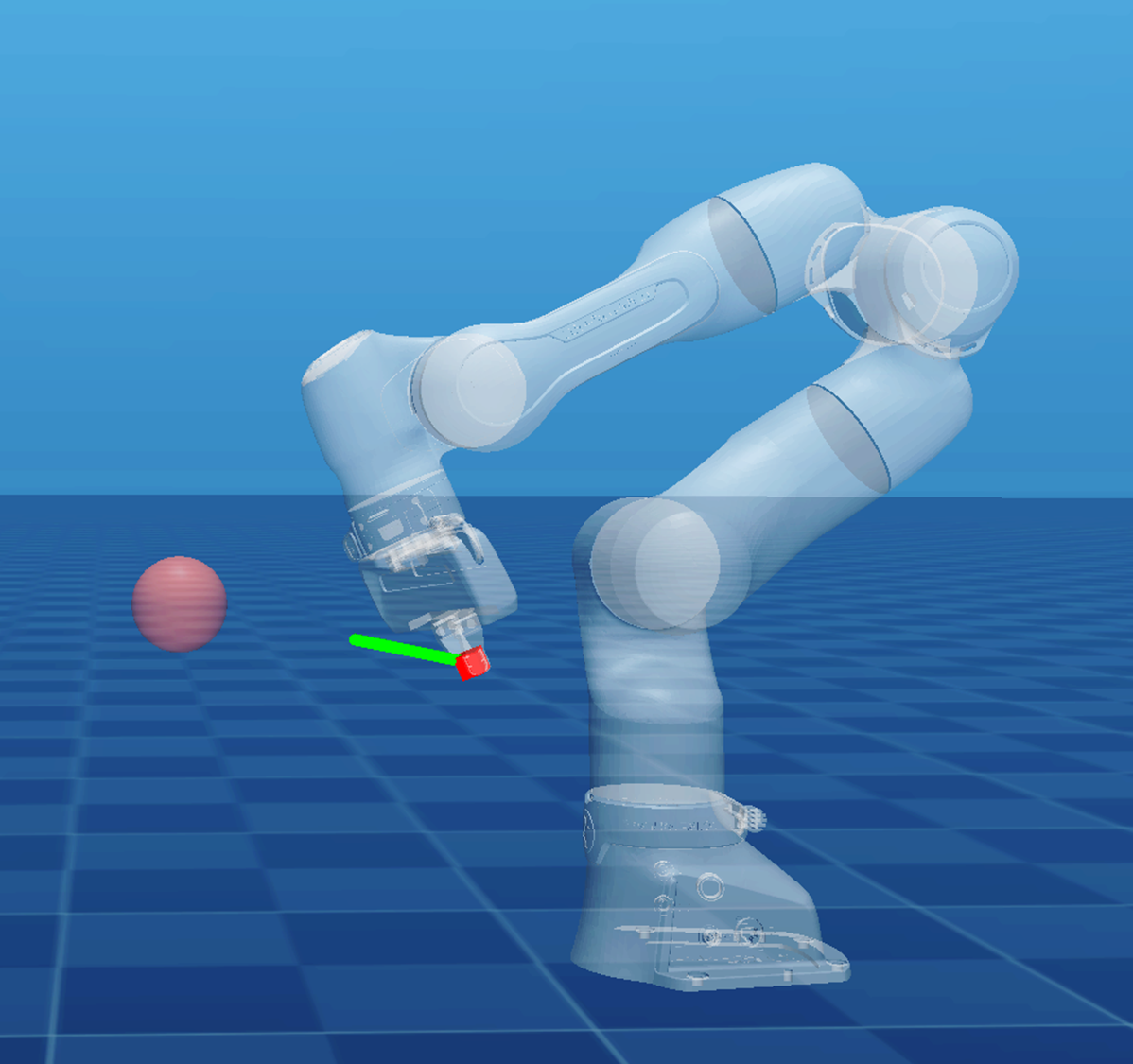}
        \caption{}
        \label{fig:ee_only}
    \end{subfigure}%
    \begin{subfigure}{.5\linewidth}
        \centering
        \includegraphics[width=.85\linewidth]{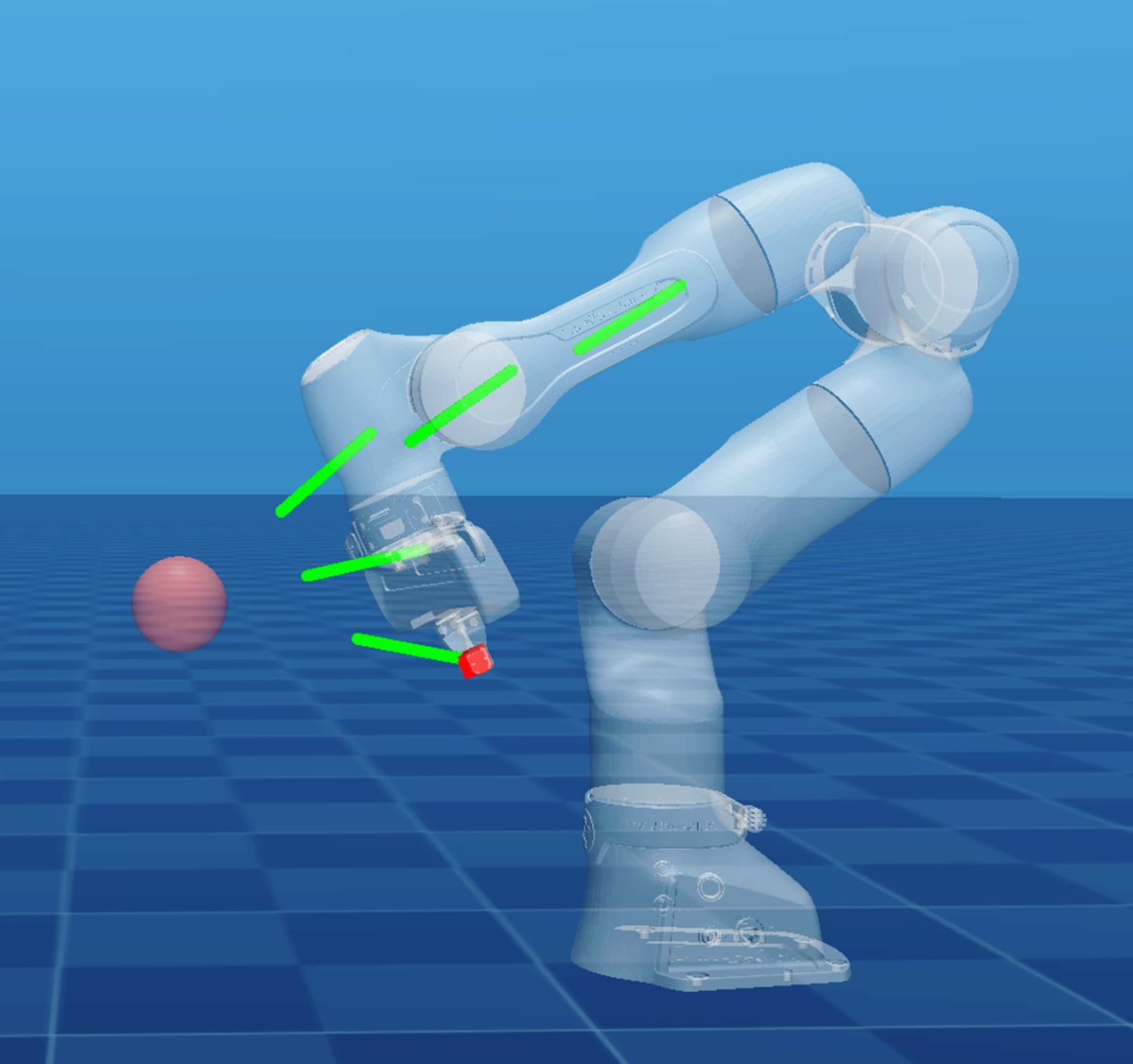}
        \caption{}
        \label{fig:full_body}
    \end{subfigure}%
    \caption{EDF sample conditions. (a) EE-setting. (b) FB-setting. }
    \label{fig:ee_fb_setting}
\end{figure}

\section{Simulation Results}
\label{sec:sim}
% We present the experimental validation of the proposed framework in both simulation and real-world environments.

% \subsection{Experimental Environment}
% \subsubsection{Simulation Experiments}
% \added{The experimental environment is simulated using MuJoCo 3.3.2.} In the simulated workspace, a FR3 model is used to train a policy to perform a reaching task, requiring the robot to navigate its EE toward a target position while maintaining safety constraints in the presence of various obstacle configurations as shown in Fig.~\ref{fig:simulation_task}.
The simulation environment is implemented in MuJoCo 3.3.2 using an FR3 model. The task requires the EE to reach a target while avoiding various obstacle configurations, as shown in Fig.~\ref{fig:simulation_task}.
\added{
All policies were trained in simulation on a laptop equipped with a 13th Gen Intel(R) Core(TM) i7-13620H CPU and an NVIDIA GeForce RTX 4060 Laptop GPU. The average policy inference time was approximately 2.2 ms per control step.
}
% In the simulation experiments, we used MuJoCo~3.3.2 and the Franka Research~3 (FR3) model. Figure~\ref{fig:simulation_task} illustrates the reaching task, where the 7-DoF arm is required to reach a target position while avoiding obstacles.

% \begin{figure}
%     \centering
%     \includegraphics[width=0.7\linewidth]{figures/simulation_task.png}
%     \caption{Simulation experimental setup}
%     \label{fig:simulation_task}
% \end{figure}

For each configuration, the environment is populated with 10 red spherical obstacles with identical radii. At the beginning of each episode, the obstacles are randomly spawned within a predefined workspace whilst maintaining a minimum inter-sphere separation to avoid overly dense or impassable configurations.
Following obstacle placement, a target position indicated by the green object, denoted as $\mathbf{p}_g$, is sampled within a feasible goal region such that the minimum clearance to all obstacles exceeds a prescribed safety margin. 
A trial is recorded as a success when the Euclidean distance between the manipulator EE and the target falls below a predefined threshold. Upon success, a new target is re-sampled within the current environment and the episode continues. The sampling ranges are summarised in Table~\ref{tab:sampling_ranges_seeds}.

\begin{table}[b]
\centering
\caption{Summary of the six simulation configurations}
\label{tab:obs_reward_configs}
\renewcommand{\arraystretch}{1.15}
\setlength{\tabcolsep}{4pt}
\small
\begin{tabular}{l p{0.58\columnwidth} c}
\toprule
Condition & Environment-related observation & APF \\
\midrule
% Baseline RL         & Obstacle centre coordinates & \texttimes \\
% RL-APF & Obstacle centre coordinates relative to EE                & \checkmark \\
% EDF-RL-EE & EDF at EE                                  & \texttimes \\
% EDF-RL-APF-EE         & EDF at EE                                  & \checkmark \\
% EDF-RL-FB & EDF at EE + $N{=}4$ body points            & \texttimes \\
% EDF-RL-APF-FB        & EDF at EE + $N{=}4$ body points            & \checkmark \\
Baseline RL          & Obstacle centre coordinates (w.r.t. world frame)                & $\times$ \\
RL-APF               & Obstacle centre coordinates (w.r.t. EE)    & $\circ$ \\
EDF-RL-EE            & EDF at EE                                  & $\times$ \\
EDF-RL-APF-EE        & EDF at EE                                  & $\circ$ \\
EDF-RL-FB            & EDF at EE + $N{=}4$ body points            & $\times$ \\
EDF-RL-APF-FB        & EDF at EE + $N{=}4$ body points            & $\circ$ \\
\bottomrule
\end{tabular}
\end{table}

\begin{table*}[!t]
\centering
\small
% \vspace*{-2mm}
\caption{Simulation results over 100 tests}
\label{tab:sim_results}
\renewcommand{\arraystretch}{1.15}
% \resizebox{\textwidth}{!}{%
\begin{tabular}{lccccc}
\toprule
\multirow{2}{*}{Condition} 
& Reaching goal 
& \multirow{2}{*}{Reaching goal (\%)} 
& Average number   
& Reaching time (s)
& Time to learning \\
& w/o collision (\%) 
&
& of collisions
& (per episode)
&  convergence (h)
\\
\midrule
Baseline RL 
& $35.0$
& $51.0$
& $27.9$
& $14.5$
& $2.23$\\

RL-APF         
& $\bm{64.0}$
& $\bm{86.0}$
& $\bm{4.10}$
& $\bm{9.38}$
& $\bm{1.92}$\\

\midrule
EDF-RL-EE 
& $62.0$
& $83.0$
& $6.20$
& $10.42$
& $1.84$\\

EDF-RL-APF-EE 
& $\bm{75.0}$
& $\bm{96.0}$
& $\bm{6.79}$
& $\bm{6.45}$
& $\bm{4.09}$\\

\midrule
EDF-RL-FB
& $56.0$
& $78.0$
& $9.82$
& $17.22$
& $2.50$\\

EDF-RL-APF-FB 
& $\bm{78.0}$
& $\bm{95.0}$
& $\bm{3.46}$
& $\bm{8.52}$
& $\bm{4.42}$\\
\bottomrule
\end{tabular}%
\end{table*}

% To investigate how the spatial extent of EDF information affects learning, we consider two observation settings (see Fig.~\ref{fig:ee_fb_setting}):
\added{To investigate the effect of EDF spatial coverage, we consider two observation settings (see Fig.~\ref{fig:ee_fb_setting}):}
\begin{enumerate}
    \item \textit{EE-only}, where the EDF distance and gradient are queried only at the EE position.
    \item \textit{FB}, where the same quantities are queried at the EE and at $N=4$ additional body points (Link~5 base and tip, Link~7 centre, and hand centre), and concatenated to the observation vector.
\end{enumerate}
Here, $N$ and the placement of the representative points are design choices; in principle, they can be increased and distributed over more links (or multiple points per link) to provide denser whole-body proximity information.

% The environment contains 10 spherical obstacles (red) with identical radii. At the beginning of each episode, the obstacles are randomly placed within a predefined workspace, while enforcing a minimum inter-sphere separation to avoid overly dense configurations. After obstacle placement, the target position (green) is sampled within a feasible goal region such that the minimum clearance to all obstacles exceeds a prescribed safety margin. A trial is considered successful when the Euclidean distance between the EE and the target falls below a threshold; upon success, a new target is re-sampled and the episode continues. The sampling ranges for obstacles and goals are summarised in Table~\ref{tab:sampling_ranges_seeds}.

% \begin{figure}[t]
%     \centering
%     \begin{minipage}[t]{0.48\linewidth}
%         \centering
%         \reflectbox{\includegraphics[width=\linewidth]{figures/ee_only.png}}
%         \caption{EE-only setting.}
%         \label{fig:ee_only}
%     \end{minipage}\hfill
%     \begin{minipage}[t]{0.48\linewidth}
%         \centering
%         \reflectbox{\includegraphics[width=\linewidth]{figures/full_body.png}}
%         \caption{FB setting.}
%         \label{fig:full_body}
%     \end{minipage}
% \end{figure}

% \begin{figure}
%     \centering
%     \includegraphics[width=0.5\linewidth]{figures/fb_edf_side.png}
%     \caption{Enter Caption}
%     \label{fig:placeholder}
% \end{figure}

\vspace{0.5em}
\noindent\textbf{Simulation configurations.}
To further differentiate the effect of EDF observations from the APF term used in reward shaping, we evaluate six configurations, as summarised in Table~\ref{tab:obs_reward_configs}. \added{All configurations share the robot motion state and desired goal position, while their environment-related observations and use of APF guidance vary.}
\begin{enumerate}
    \item \textbf{Baseline RL}: no EDF-derived proximity cues are provided. Instead, the observation contains global obstacle information, given as centres of all $10$ spherical obstacles w.r.t. the world frame.
    % \item \textbf{RL-APF}: identical observations to \textbf{No EDF}, but with reward shaping based on APF.
    \item \textbf{RL-APF}: The observation contains global obstacle information, given as the relative vectors from the EE to the centres of all $10$ spherical obstacles. The reward shaping is based on the simulator calculated APF.
    % identical observations to \textbf{Baseline RL}, but with reward shaping based on simulator calculated APF.
    \item \textbf{EDF-RL-EE}: \textit{EE-only} EDF observations.
    \item \textbf{EDF-RL-APF-EE}: \textit{EE-only} EDF observations, with the APF using EDF distance and gradient term.
    \item \textbf{EDF-RL-FB}: \textit{FB} EDF observations.
    \item \textbf{EDF-RL-APF-FB}: \textit{FB} EDF observations, with reward shaping based on APF.
 
\end{enumerate}

% Configurations 1) and 2) assume access to the obstacle centre positions, and therefore the APF term can only be computed inside the simulator. In contrast, configurations 4)--6) use the distance and gradient provided by the EDF to construct the APF term; these quantities can be computed by IDMP and are thus applicable to real-world settings. In practice, ground-truth obstacle positions are not directly available without additional instrumentation, making the baselines in 1) and 2) unrealistic outside simulation.

% Configurations 1) and 2) assume access to the obstacle centre positions, and therefore the APF term can only be computed inside the simulator.
Configurations 1) and 2) assume access to ground-truth obstacle centre positions and are therefore limited to simulation. In contrast, configurations 3)--6) use the distance and gradient provided by the EDF as observations, while configurations 4) and 6) additionally use these quantities to construct the APF term; these quantities can be computed by IDMP and are thus applicable to real-world settings. In practice, ground-truth obstacle positions are not directly available without additional instrumentation, making the baselines in 1) and 2) unrealistic outside simulation.

%We now analyse the experimental results.
% Need to explain how to get all the data
% Simulation numbers: 100
% Success ratio: after 2000 steps...

The results in Table~\ref{tab:sim_results} are computed from 100 simulation trials as follows:
\begin{itemize}
  \item \textbf{Reaching goal (\%):} the percentage of episodes in which the EE reaches the goal (with or without collision) within 2000 steps.
  \item \textbf{Reaching goal w/o collision (\%):} the percentage of episodes in which the goal is reached without any collision during the episode.
  \item \textbf{Average number of collisions:} the average number of collision events per episode. A collision event is counted when the robot touches any obstacle sphere; continuous contact is counted as one event, and a new event is counted only after the robot separates and touches again.
  \item \textbf{Reaching time (s):} for successful episodes, the time from episode start until the goal is reached, averaged over successful episodes.
  \item \textbf{Time to learning convergence (h):} the elapsed time until the learning curve is judged to have converged based on a predefined threshold.
\end{itemize}

As shown in Table \ref{tab:sim_results}, integrating EDF into RL-APF-based strategies significantly improves performance compared to Baseline RL and RL-APF methods. Specifically, EDF-RL-APF-EE and EDF-RL-APF-FB generally show better goal-reaching success, fewer collisions, and faster reaching times. However, there are instances where they do not consistently outperform other strategies.

% For example, EDF-RL-APF-EE performs better than RL-APF in goal-reaching success and time, but it experiences slightly more collisions in some cases.

% For example, EDF-RL-APF-EE outperforms RL-APF in both goal-reaching success rate and reaching time, but it can incur slightly more collisions in some episodes when executing recovery manoeuvres.

% For example, EDF-RL-APF-EE outperforms RL-APF in both
% \added{the goal-reaching}
% success rate and
% \added{reach }
% time, but it can incur slightly more collisions in some episodes, when
% \added{performing}
% recovery
% \added{maneuvers}.

For example, EDF-RL-APF-EE achieves a higher \added{collision-free goal-reaching success rate} and a \added{shorter reaching time} than RL-APF, although it can incur slightly more collisions in some episodes when \added{performing recovery maneuvers}.

% EDF-RL-APF-FB, which combines EDF with feedback control, reduces collisions and improves goal-reaching efficiency but may not always be the fastest.

\added{EDF-RL-APF-FB achieves the highest collision-free success rate and the lowest average number of collisions, although EDF-RL-APF-EE achieves a shorter reaching time.}

% When EDF is used without APF (EDF-RL-EE and EDF-RL-FB), the results are mixed. EDF-RL-EE shows improved goal-reaching success and reduced time but struggles with collision avoidance compared to EDF-RL-APF methods. Similarly, EDF-RL-FB benefits from feedback control to reduce collisions but does not match the efficiency of EDF-APF combinations.

\added{While EDF-RL-EE and EDF-RL-FB improve collision-free success over the Baseline RL condition, EDF-RL-APF-EE and EDF-RL-APF-FB achieve higher collision-free success rates and shorter reaching times.}

Overall, EDF-based strategies, especially when combined with RL-APF, offer improvements in safety (fewer collisions) and efficiency (faster goal-reaching), though performance varies depending on the evaluation criteria.

\section{Real-World Experiments}

In the real-world experiments, we performed an obstacle-avoidance reaching task, where the objective is to move the EE to a target position while avoiding obstacles. \added{Note that the RL policy was trained only in simulation, as described above, and transferred zero-shot to the real robot.}
% Note that the RL policy is trained only in simulation as per the section above; thus, we performed a zero-shot sim-to-real transfer. 

\begin{figure}[t]
    \centering
    \begin{subfigure}{.48\linewidth}
        \centering
        \includegraphics[width=.95\linewidth]{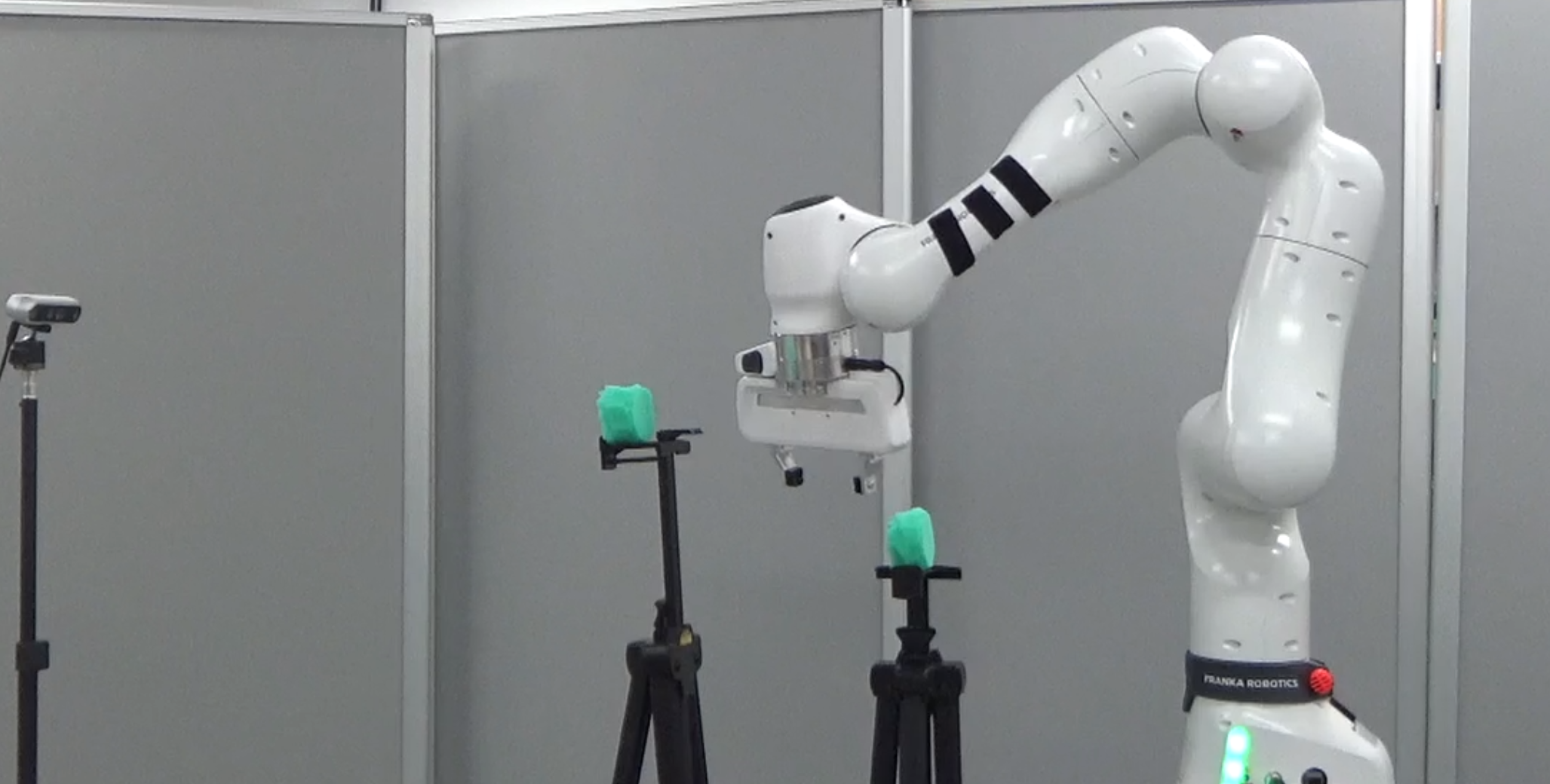}
        \caption{}
        \label{fig:realworld1}
    \end{subfigure}%
    \begin{subfigure}{.48\linewidth}
        \centering
        \includegraphics[width=.95\linewidth]{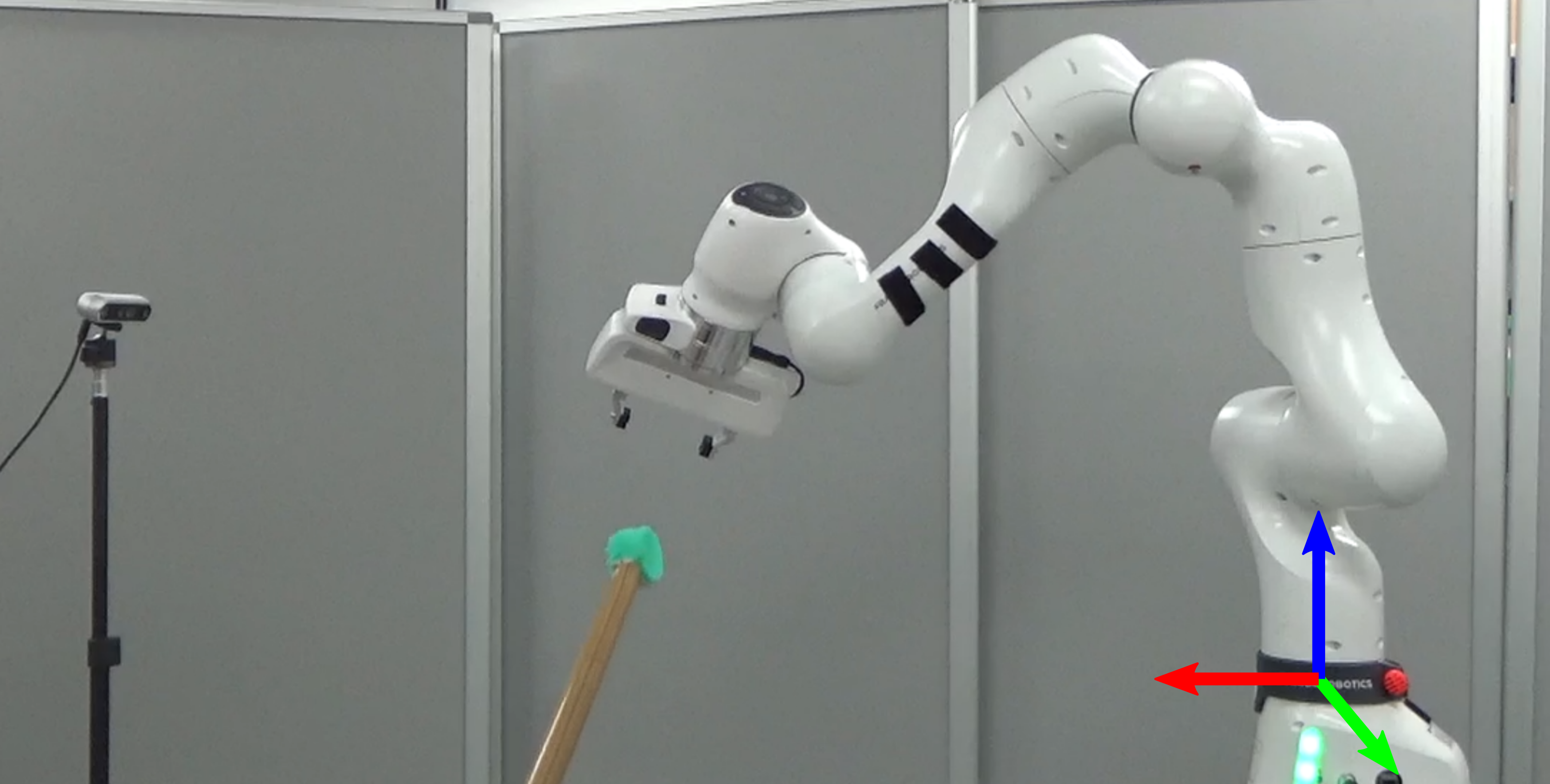}
        \caption{}
        \label{fig:realworld2}
    \end{subfigure}%
    \caption{Real-world experimental setups: (a) two obstacles in predefined positions; (b) a single dynamic obstacle introduced during execution.}
    \label{fig:realworld_exp}
    \vspace*{-3mm}
\end{figure}

% We evaluated two scenarios: (i) a setup with two obstacles arranged in predefined positions to confirm reachability,
We evaluated two scenarios: (i) a setup with two static obstacles,
as shown in Fig.~\ref{fig:realworld_exp} (a), and (ii) a scenario where the task starts without obstacles and a single obstacle is introduced into the environment during the execution of the reaching task, as illustrated in Fig.~\ref{fig:realworld_exp} (b), to assess avoidance and goal-reaching performance under the presence of a dynamic obstacle.

For environmental sensing, an Intel RealSense D455 RGB-D camera was used to acquire depth measurements. The point cloud was reconstructed from the depth data, and the EDF distance and gradient vector at an arbitrary query position were inferred using the IDMP framework. The distance and gradients at the requested positions were provided to the agent in the same observation format as in the simulation experiments.

\added{
In the real-world experiments, the learned policy was executed on the same laptop. The policy outputs were converted into joint-position targets based on the current joint configuration and sent to the Franka controller.
}

As in simulation, the Baseline RL and RL-APF assume knowledge of the obstacle ground-truth positions. 
%This is not considered for the real experiments since pre-mapping and dedicated instrumentation is required for the dynamic obstacles, which is not required to test the proposed policies.
We did not consider these cases in the real experiment, as they require a pre-mapping stage and dedicated instrumentation for the dynamic obstacle, which is not required for the proposed policies. 
Therefore, we evaluated only four EDF-based configurations: EDF-RL-EE, EDF-RL-APF-EE, EDF-RL-FB and EDF-RL-APF-FB with static obstacles.

\begin{figure*}[!t]
    \centering
    \setlength{\tabcolsep}{0pt}   % horizontal gap between columns
    \renewcommand{\arraystretch}{1}

    \begin{tabular}{ccccc}
        % Row 1: IDMP
        \includegraphics[width=0.195\textwidth]{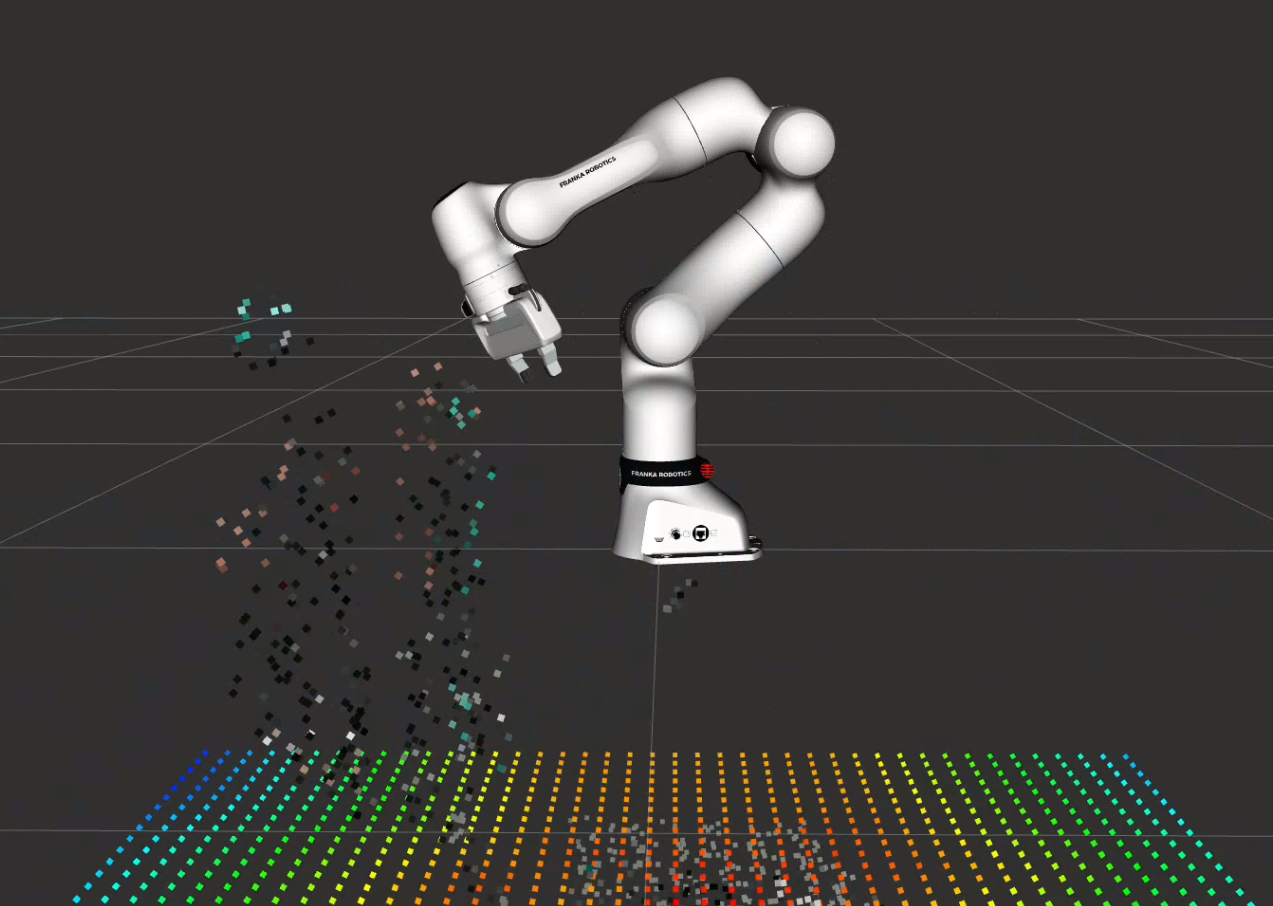} &
        \includegraphics[width=0.195\textwidth]{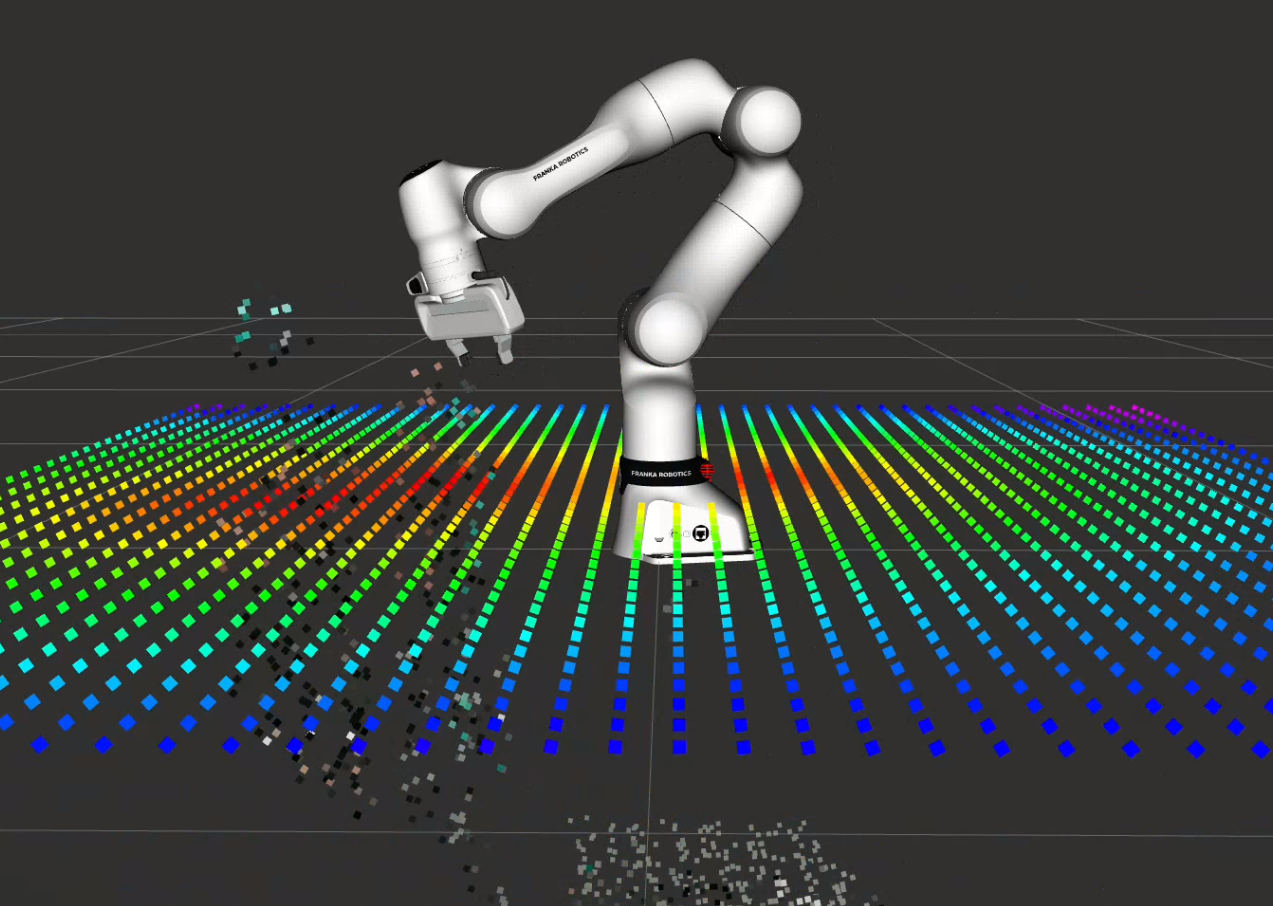} &
        \includegraphics[width=0.195\textwidth]{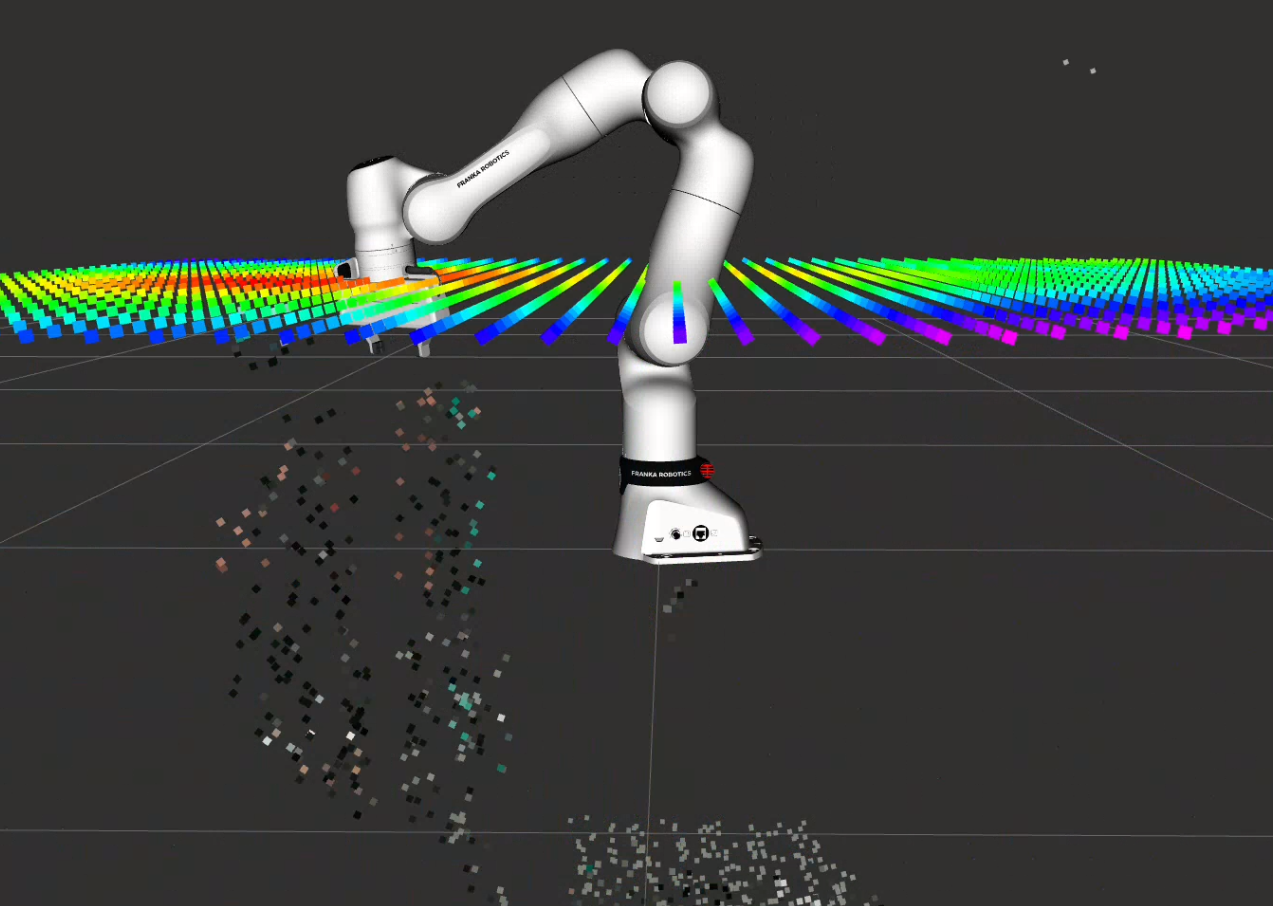} &
        \includegraphics[width=0.195\textwidth]{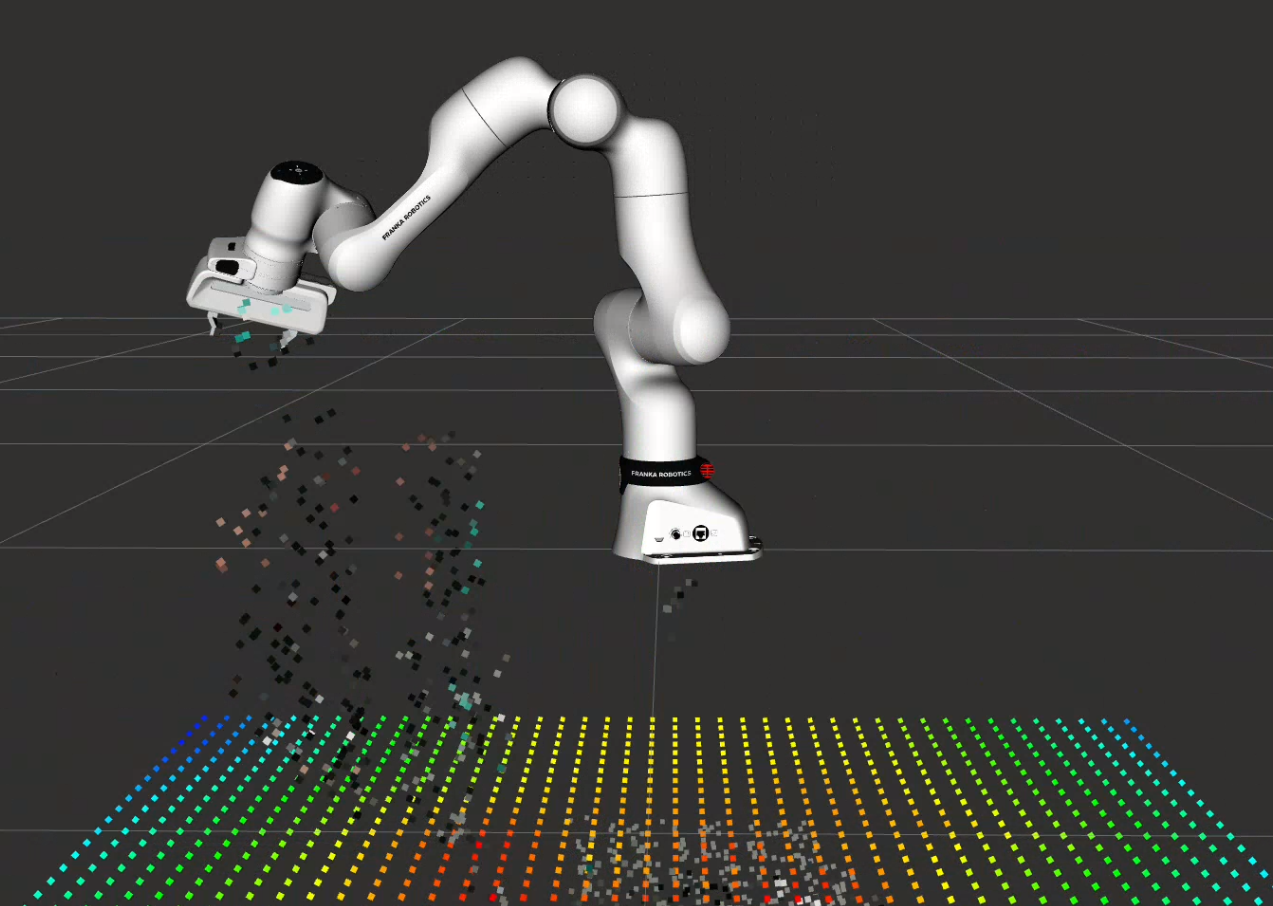} &
        \includegraphics[width=0.195\textwidth]{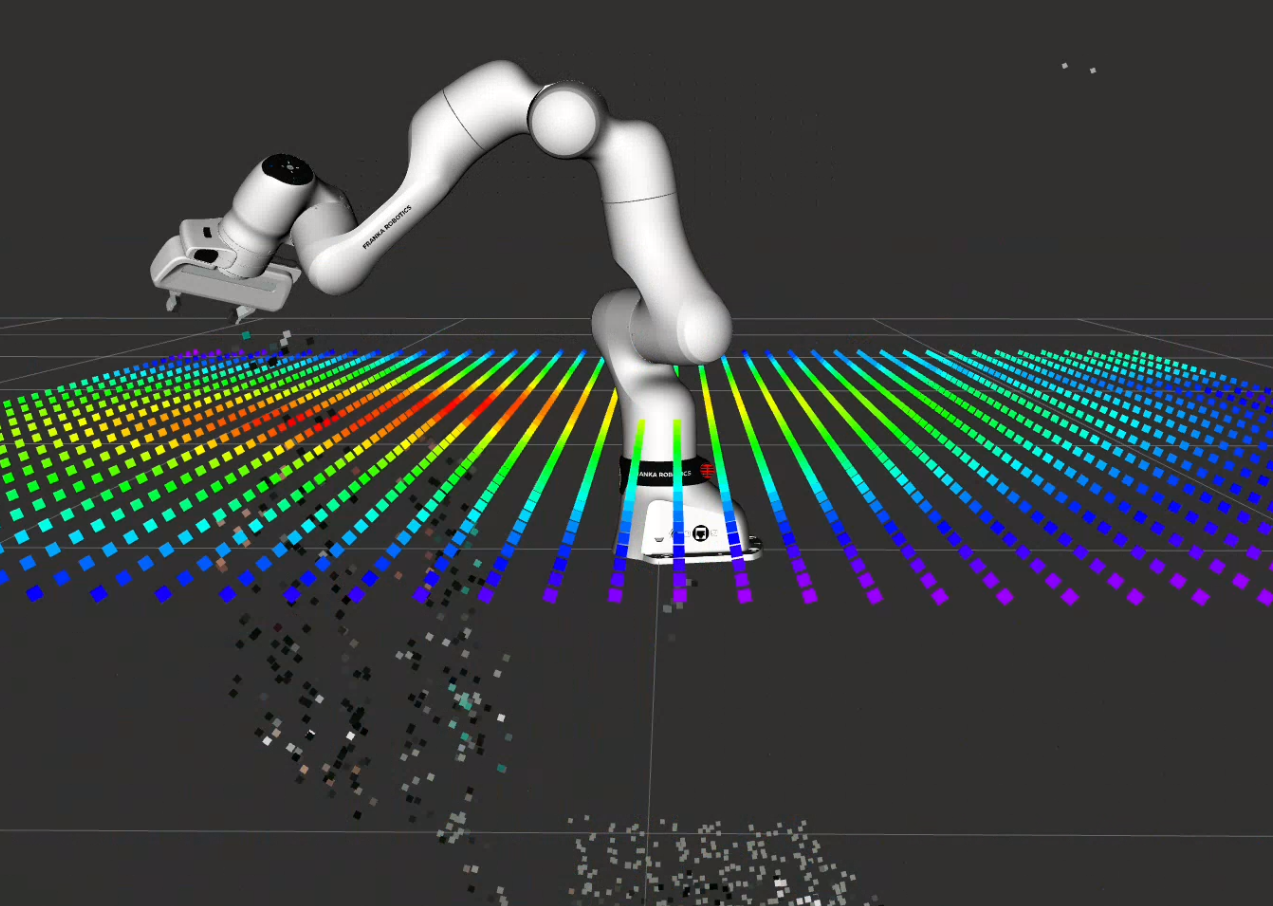} \\[0mm]

        % Row 2: Simulation
        \includegraphics[width=0.195\textwidth]{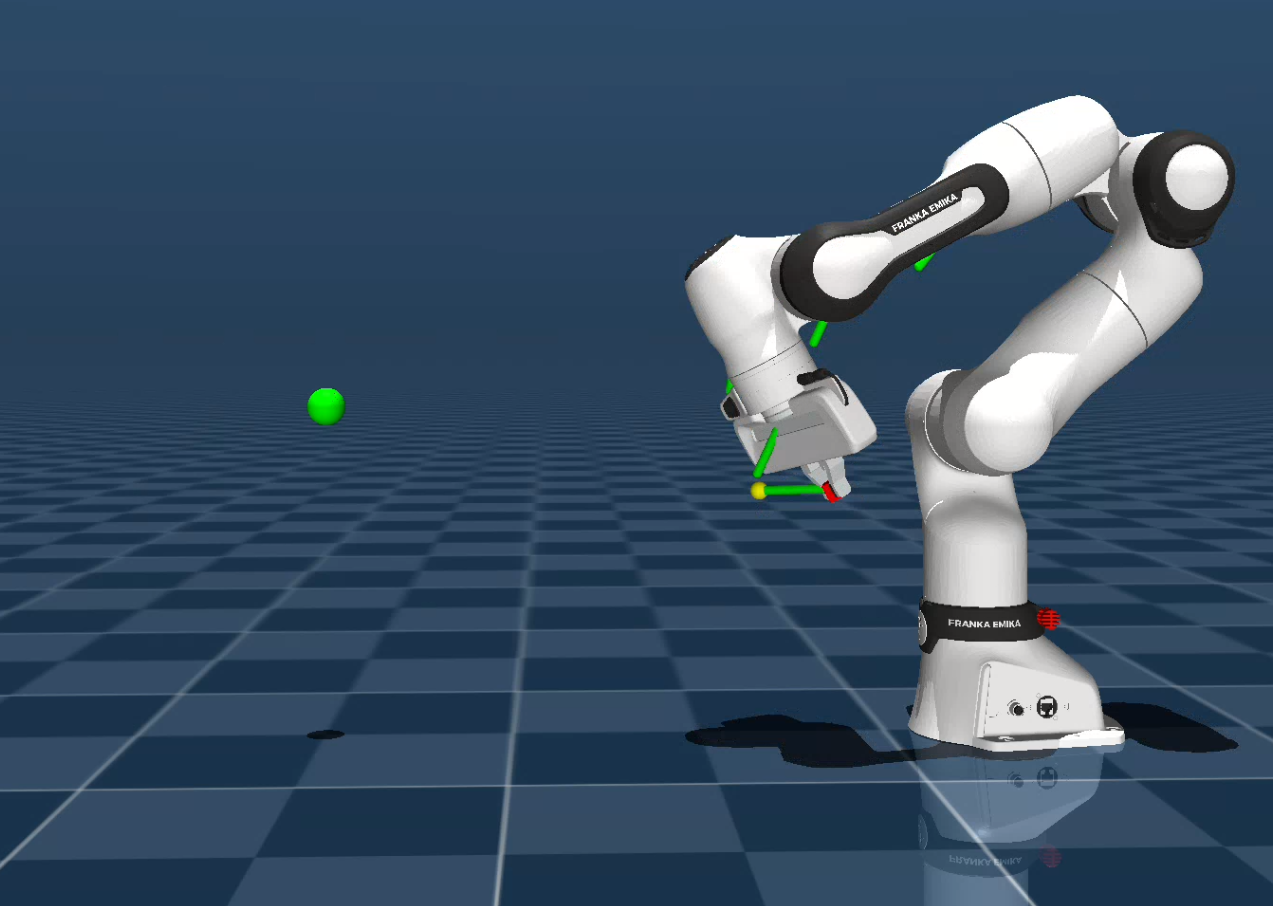} &
        \includegraphics[width=0.195\textwidth]{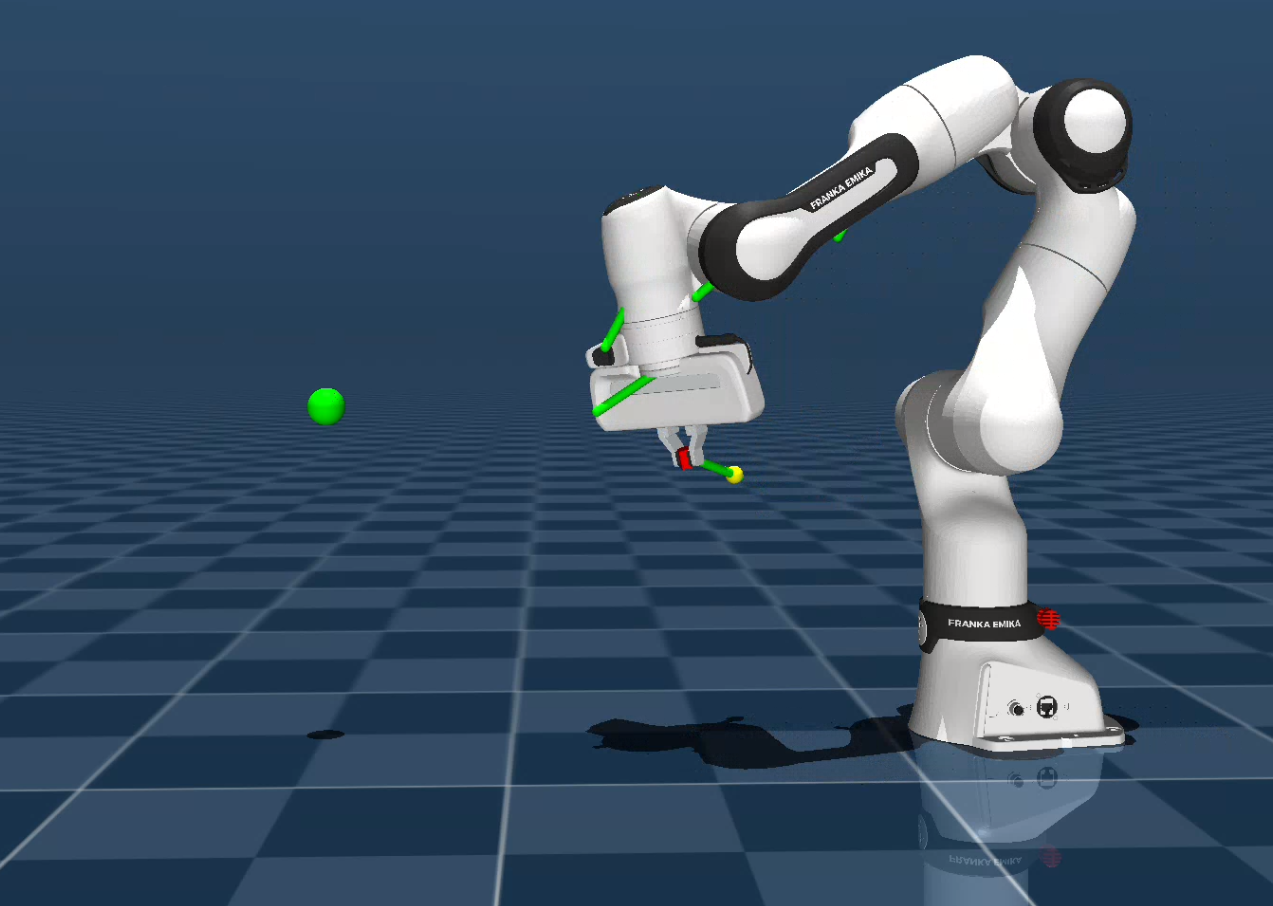} &
        \includegraphics[width=0.195\textwidth]{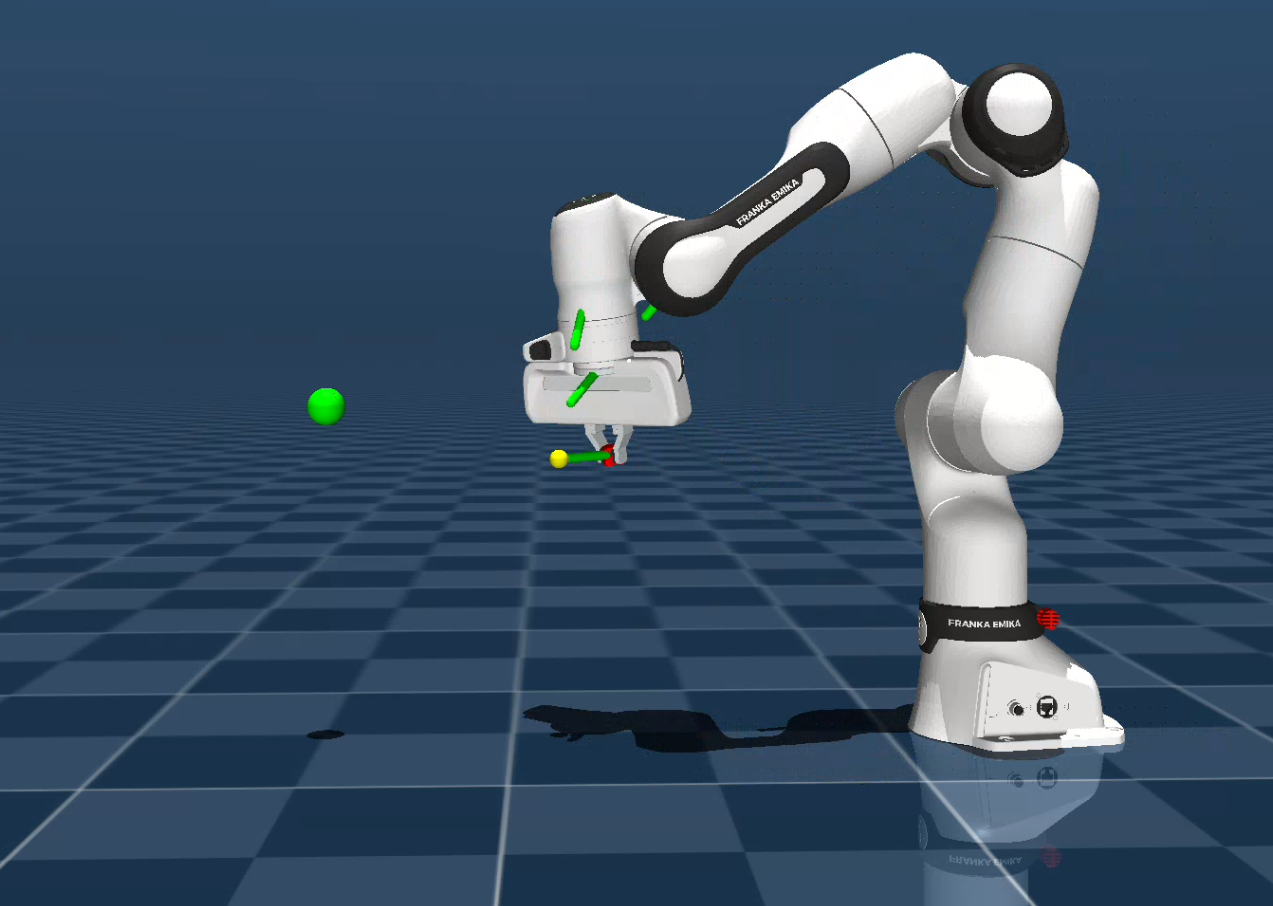} &
        \includegraphics[width=0.195\textwidth]{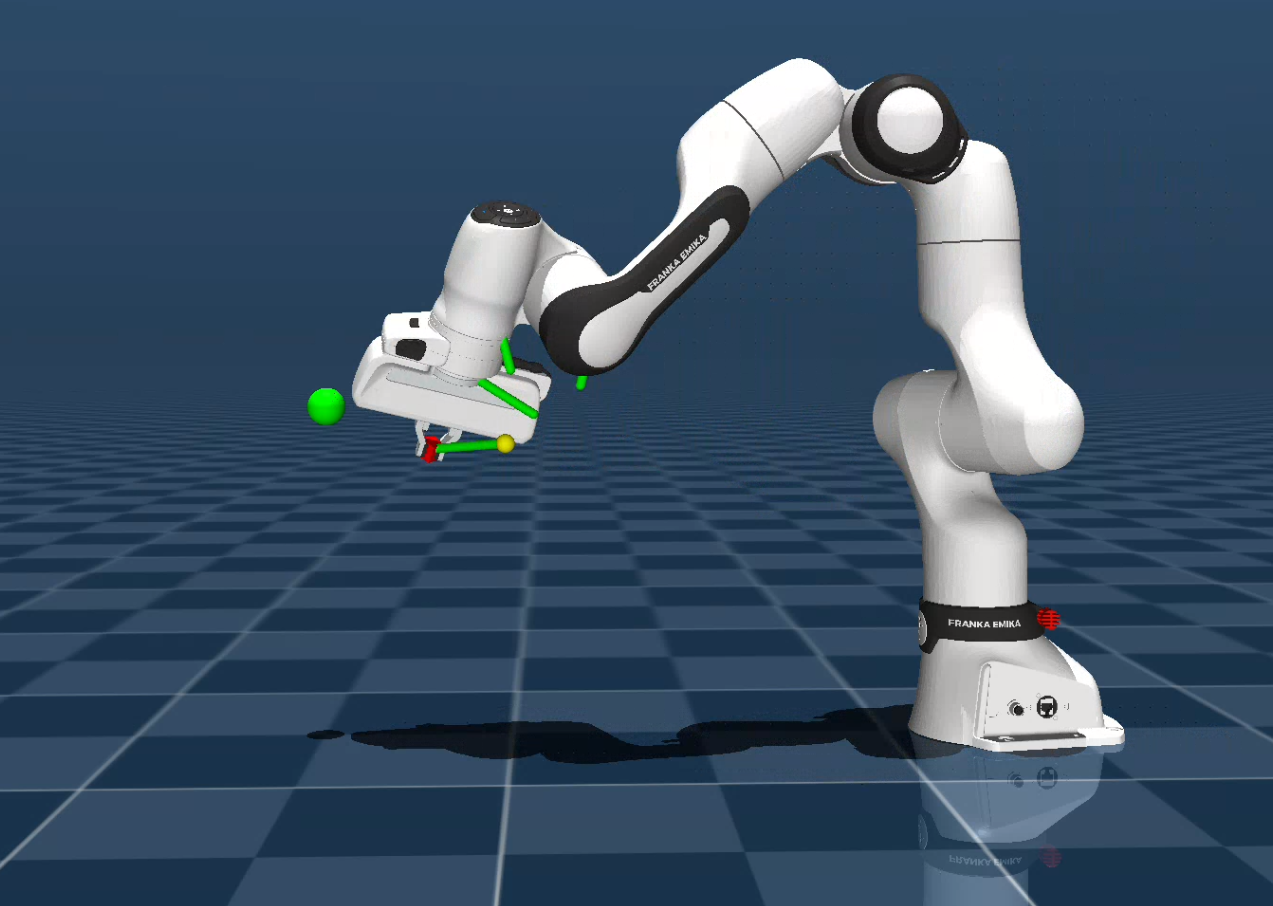} &
        \includegraphics[width=0.195\textwidth]{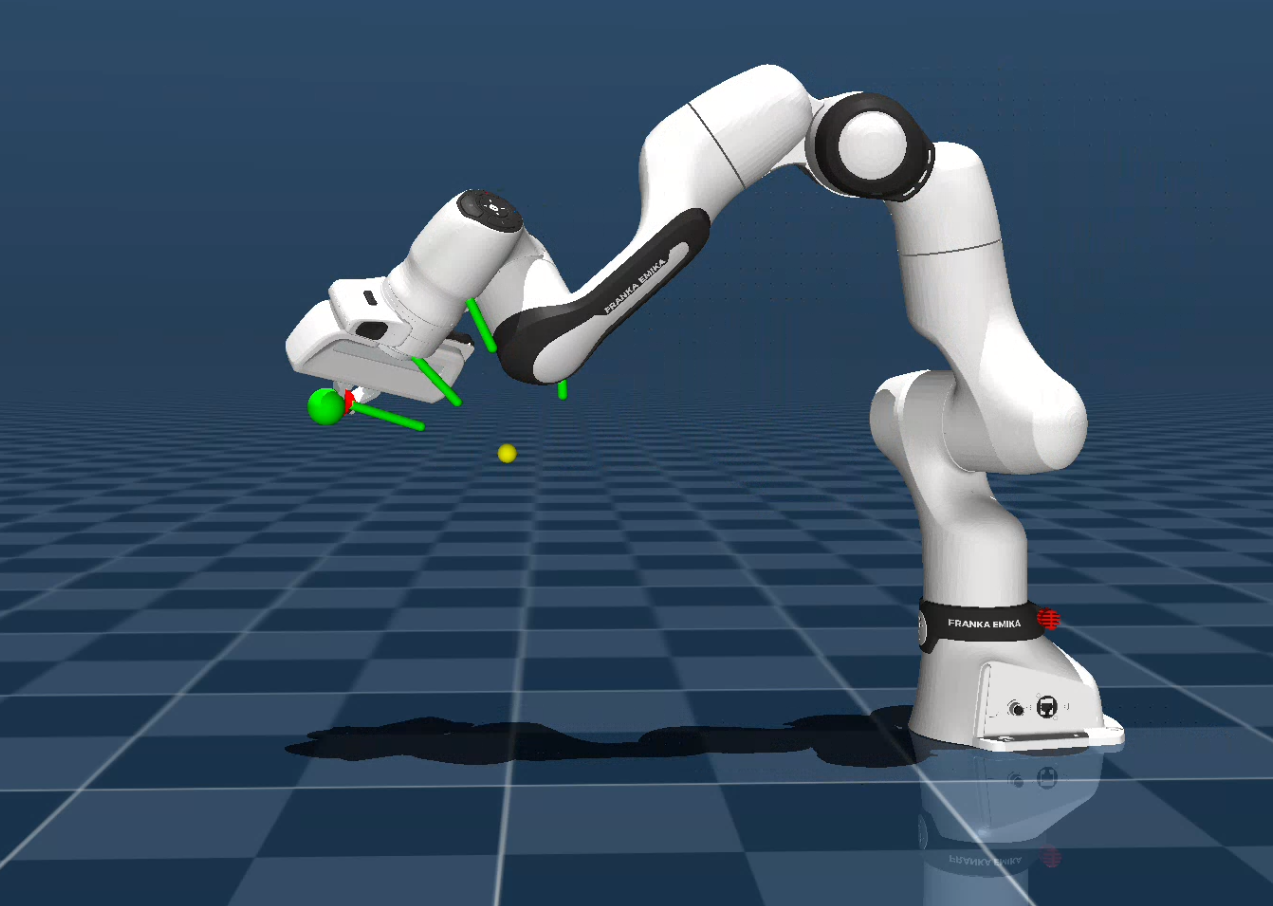} \\[0mm]

        % Row 3: Real robot
        \includegraphics[width=0.195\textwidth]{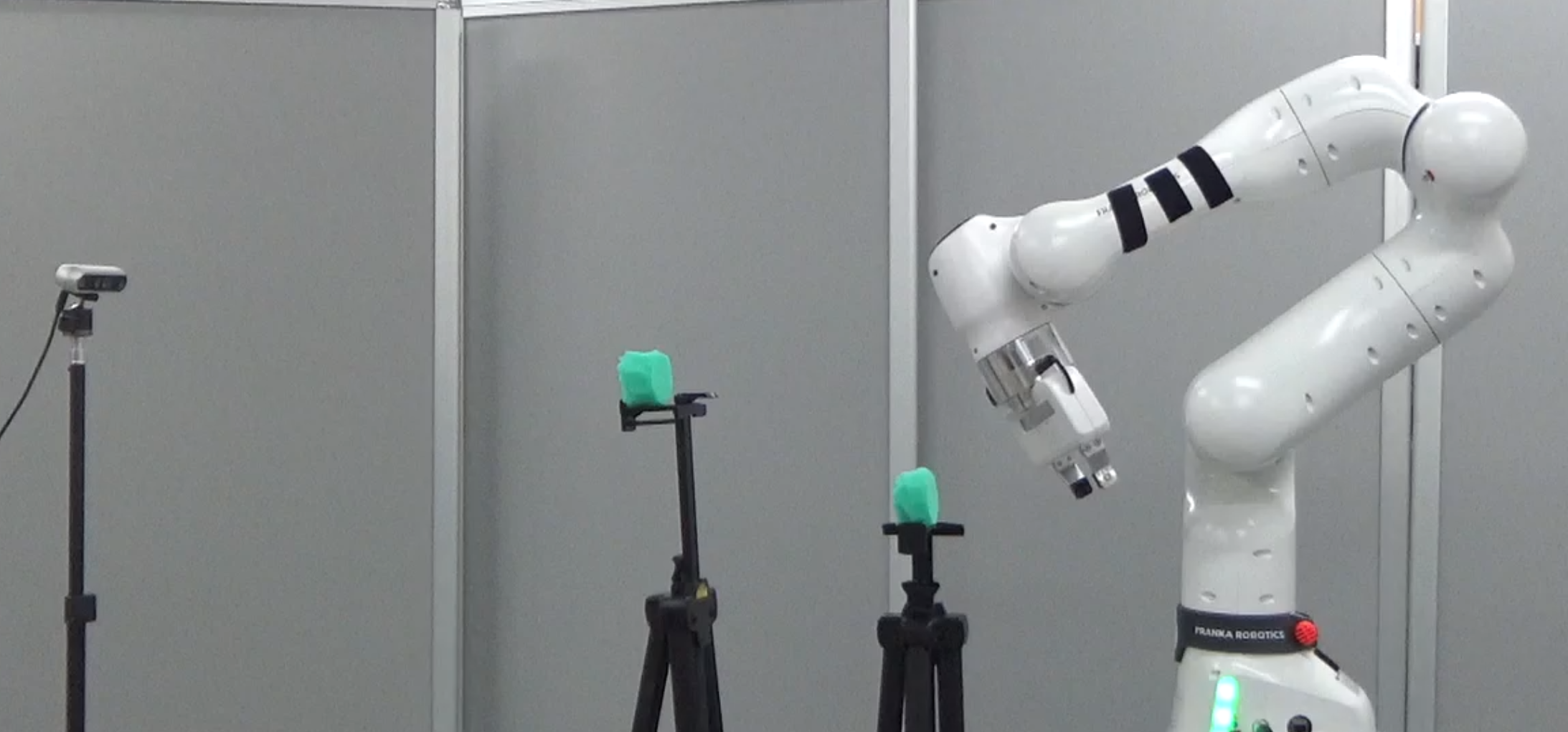} &
        \includegraphics[width=0.195\textwidth]{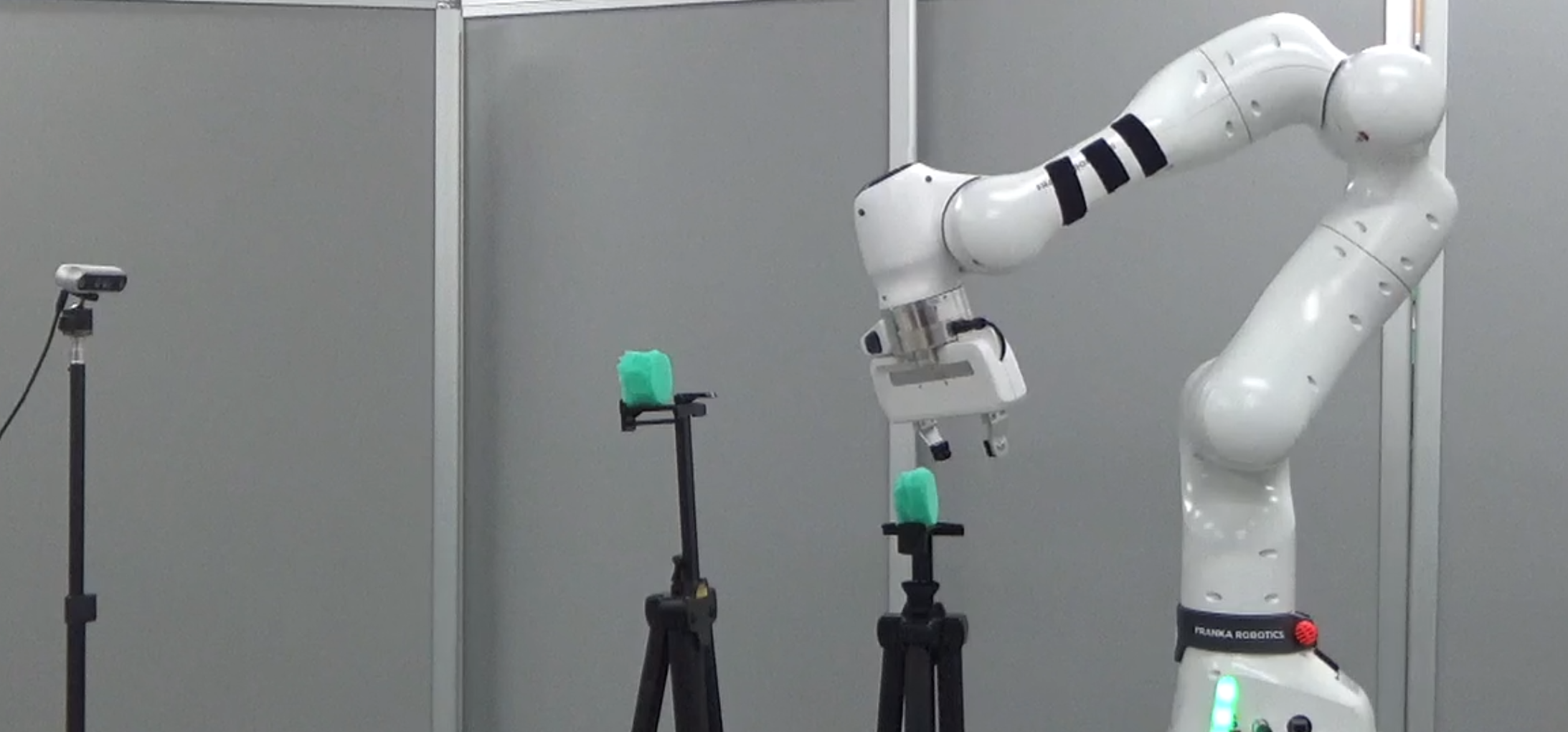} &
        \includegraphics[width=0.195\textwidth]{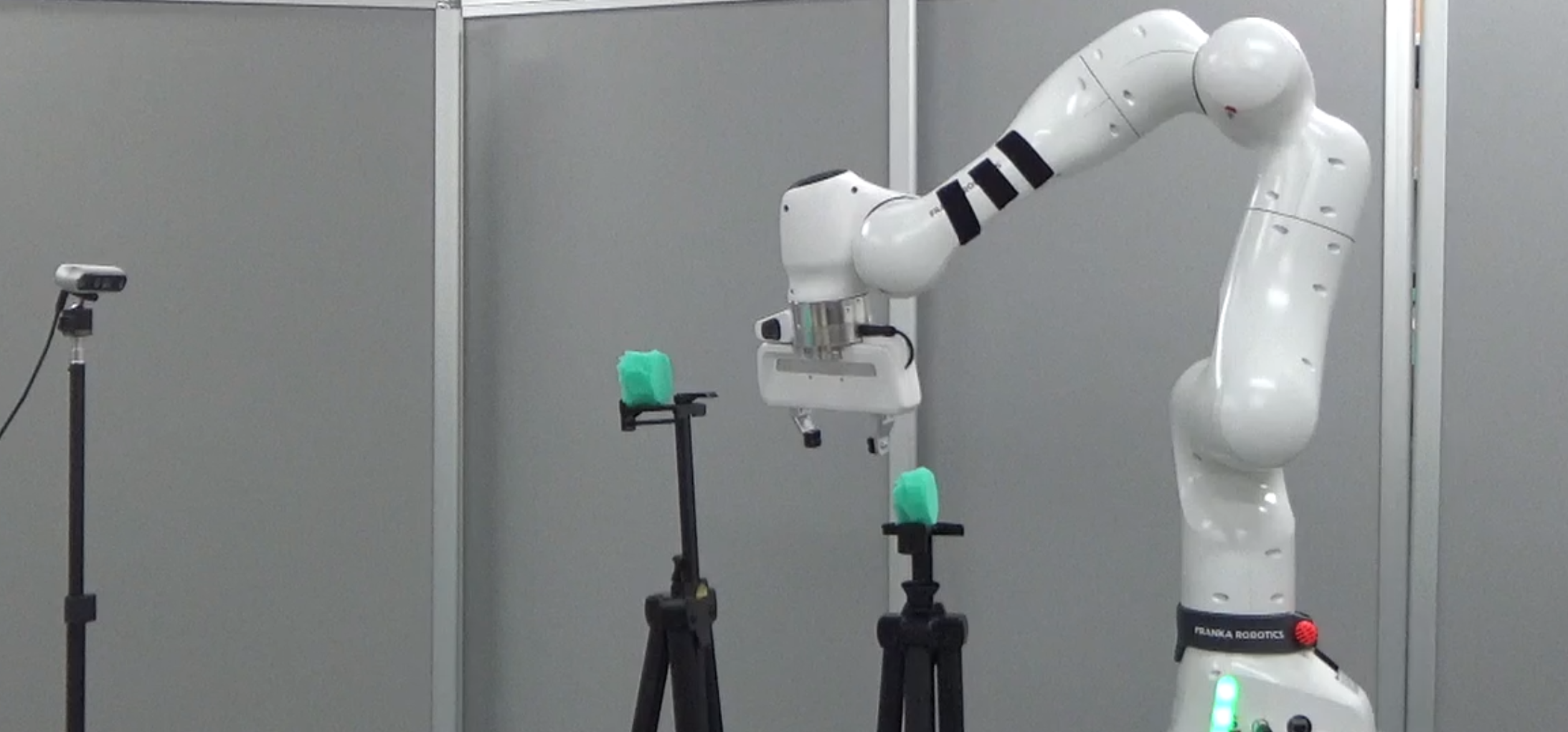} &
        \includegraphics[width=0.195\textwidth]{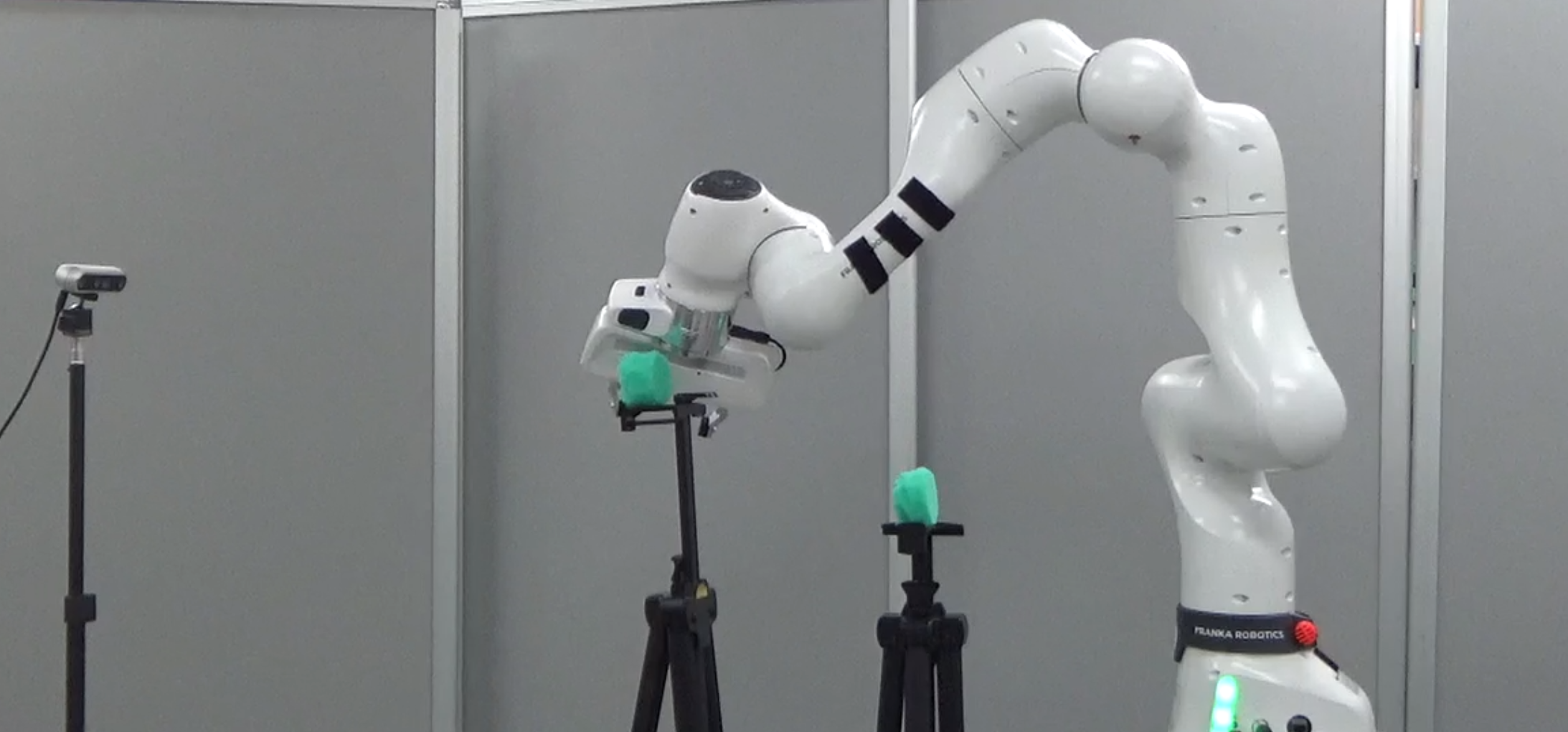} &
        \includegraphics[width=0.195\textwidth]{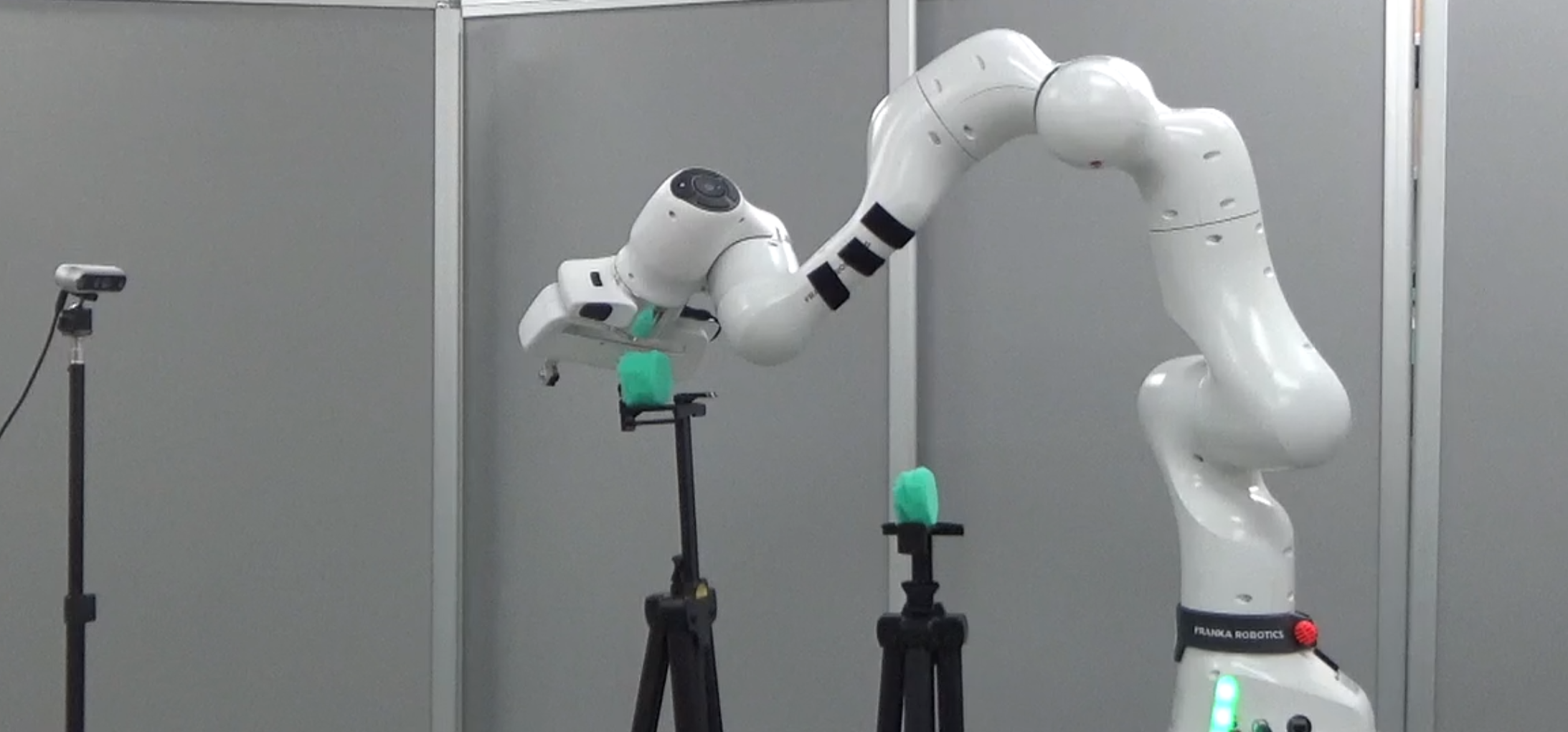} \\[-1mm]

        \multicolumn{1}{c}{\scriptsize (1)} &
        \multicolumn{1}{c}{\scriptsize (2)} &
        \multicolumn{1}{c}{\scriptsize (3)} &
        \multicolumn{1}{c}{\scriptsize (4)} &
        \multicolumn{1}{c}{\scriptsize (5)} \\
    \end{tabular}

    \caption{Time sequence (left to right) of the EDF-RL-APF-FB policy on the real robot, successfully avoiding two obstacles under the fixed-position condition, with corresponding rows (top: IDMP, middle: simulation for visual reference, bottom: real robot).}
    \label{fig:fb_3x5_idmp_simu_real}
\end{figure*}

% \begin{figure}[t]
%     \centering
%     \begin{subfigure}{.48\linewidth}
%         \centering
%         \includegraphics[width=.95\linewidth]{fixed.png}
%         \caption{}
%         \label{fig:setup_fixed}
%     \end{subfigure}%
%     \begin{subfigure}{.48\linewidth}
%         \centering
%         \includegraphics[width=.95\linewidth]{interactive.png}
%         \caption{}
%         \label{fig:setup_interactive}
%     \end{subfigure}%
%     \caption{Real-world experimental setups: (a) two obstacles in predefined positions; (b) a single dynamic obstacle introduced during execution.}
%     \label{fig:realworld_setups}
% \end{figure}

In Table~\ref{tab:real_results_rates}, \textbf{Reaching goal w/o collision} denotes the percentage of trials in which the robot reached the goal without making contact with any obstacle.
In the experiments, robot-environment collisions were detected for the robot arm including EE, 
and each trial was terminated immediately upon any contact. 

% In the event of failure, occasionally occlusions occurred so that the obstacle hid parts of the arm. This reduced the reliability of the EDF computed from the sampled points, making it harder for the policy to avoid obstacles in the occluded regions.

% -------------------- Real-world results table --------------------
\begin{table}[t]
\centering
\caption{Real-world results for the fixed-obstacle condition over 10 tests}
\label{tab:real_results_rates}
\renewcommand{\arraystretch}{1.05}
\setlength{\tabcolsep}{3pt}
\small
\begin{tabular}{lc}
\toprule
Condition & Reaching goal w/o collision (\%) \\
\midrule
EDF-RL-EE      & 60\\
EDF-RL-APF-EE  & 40\\
EDF-RL-FB      & 70\\
EDF-RL-APF-FB  & 70\\
\bottomrule
\end{tabular}
\vspace*{-3mm}
\end{table}

% Fig.~\ref{fig:fb_3x5_idmp_simu_real} shows a left-to-right time sequence when executing the EDF-RL-APF-FB policy on the real robot, successfully avoiding two obstacles under the fixed-position condition; the rows correspond to (top) IDMP, (middle) robot display in the simulation environment for visual reference, and (bottom) the real robot.

Fig.~\ref{fig:fb_3x5_idmp_simu_real} shows a left-to-right time sequence of the EDF-RL-APF-FB policy successfully avoiding two fixed obstacles on the real robot. The rows show (top) IDMP, (middle) the corresponding simulation visualisation, and (bottom) the real robot.

In the interactive (dynamic obstacle) experiment, we evaluated the EDF-RL-APF-FB policy by comparing the resulting trajectories under three conditions:
\begin{enumerate}
  \item No obstacle.
  \item Lateral dynamic interference, where a moving obstacle approaches the robot's path along the \textit{xy}-plane, defined by the coordinate frame in Fig.~\ref{fig:realworld_exp} (b).
  % \item Vertical dynamic interference, where a moving obstacle approaches the robot's path along the \textit{x}-axis from below.
  \item Vertical dynamic interference, where an obstacle approaches the robot's path from below along the \textit{z}-axis.
\end{enumerate}

\begin{figure*}[t]
    \centering
    \begin{minipage}[t]{0.23\textwidth}
        \centering
        \includegraphics[width=\linewidth]{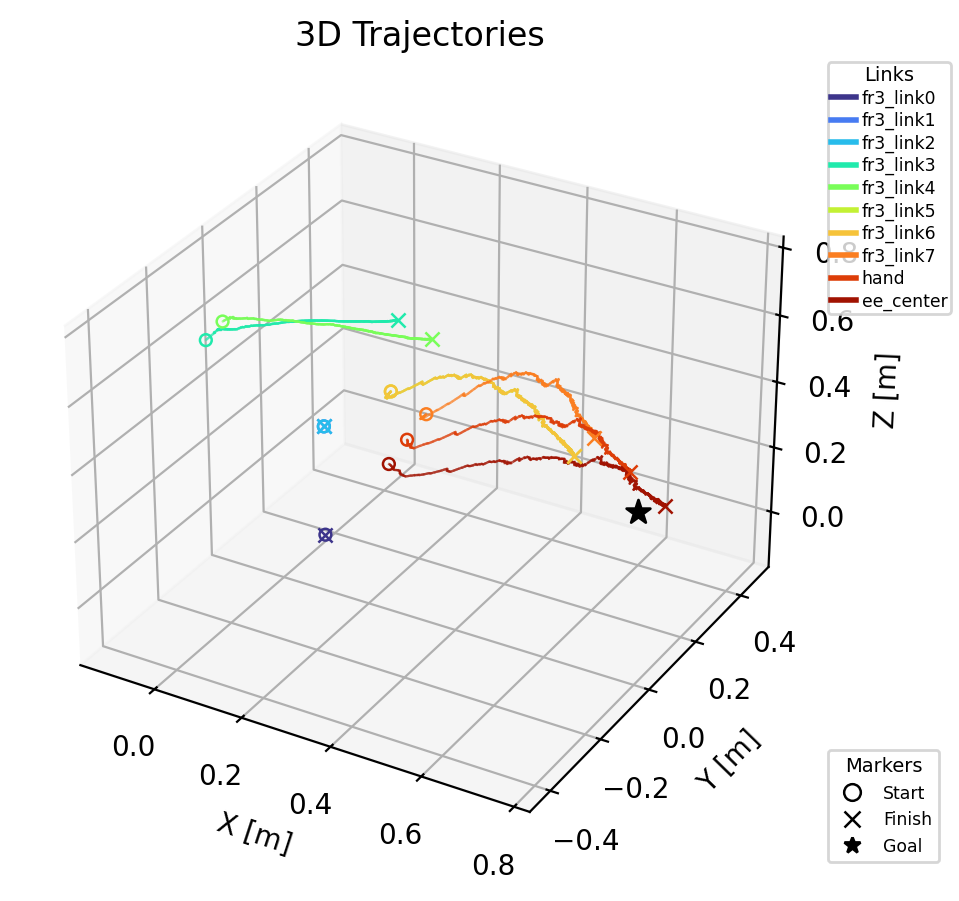}
        \caption*{\textbf{No obstacle}}
    \end{minipage}\hfill
    \begin{minipage}[t]{0.23\textwidth}
        \centering
        \includegraphics[width=\linewidth]{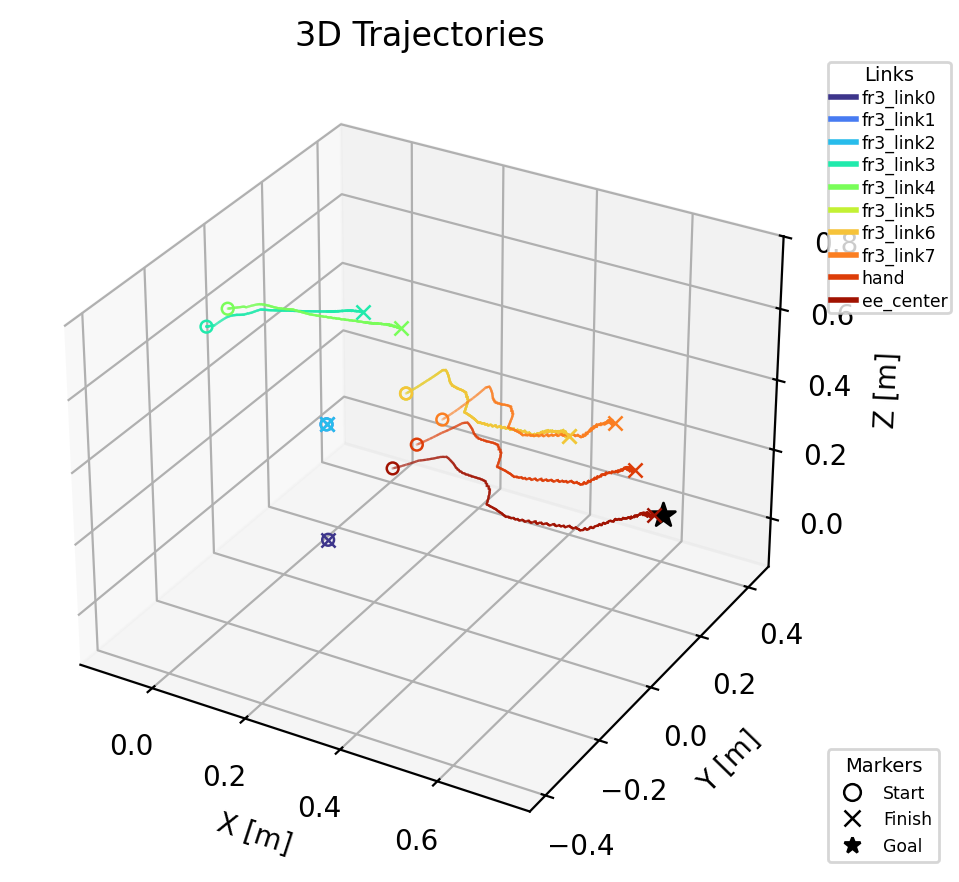}
        \caption*{\textbf{Fixed obstacles}}
    \end{minipage}\hfill
    \begin{minipage}[t]{0.23\textwidth}
        \centering
        \includegraphics[width=\linewidth]{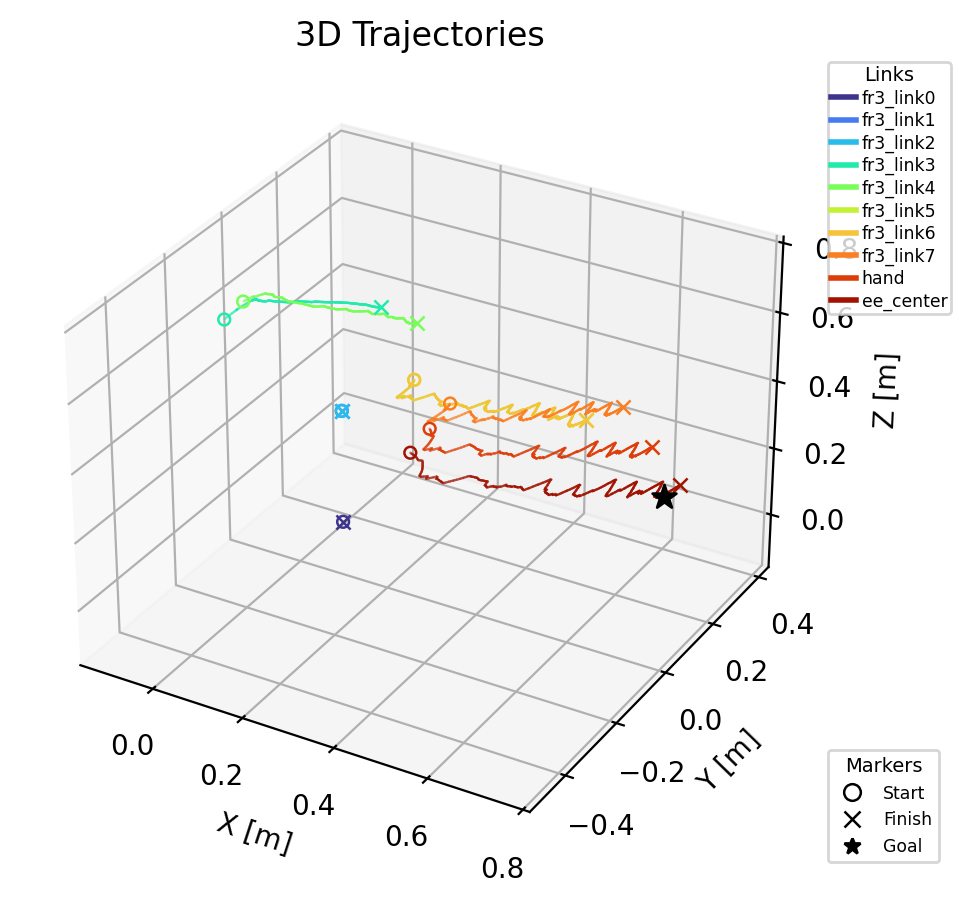}
        \caption*{\textbf{Lateral dynamic object}}
    \end{minipage}\hfill
    \begin{minipage}[t]{0.23\textwidth}
        \centering
        \includegraphics[width=\linewidth]{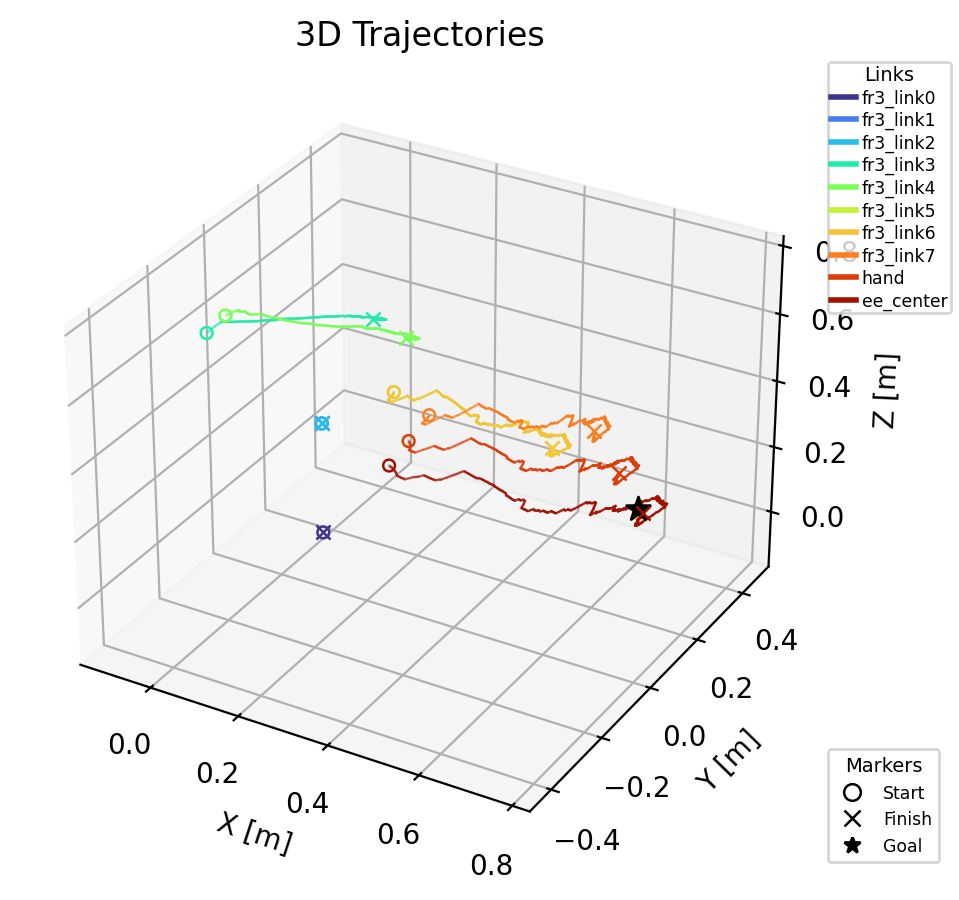}
        \caption*{\textbf{Vertical dynamic object}}
    \end{minipage}

    \caption{Trajectories of the EDF-RL-APF-FB policy under four conditions: no obstacle, fixed obstacles, lateral dynamic interference, and vertical dynamic interference.}
    \label{fig:interactive_trajectories}
    \vspace*{-3mm}
\end{figure*}

Fig.~\ref{fig:interactive_trajectories} shows trajectories generated by the EDF-RL-APF-FB policy under four conditions: no obstacle, fixed obstacles, lateral dynamic interference, and vertical dynamic interference. Compared with the no-obstacle case, the presence of obstacles induces clear deviations in the trajectory, indicating effective avoidance behaviour in response to both static and dynamic interference. During the interactive experiments, the obstacle exhibits periodic oscillatory motion as it is introduced to interfere with the robot path. In response, the EDF-RL-APF-FB policy generates corresponding reactive trajectories for the distal links (Links 6 and 7), the hand, and the end-effector centre, as illustrated in Fig.~\ref{fig:interactive_trajectories}.
%In the interactive conditions, where the obstacle is introduced back and forth during execution, the robot continually reacts to the changing obstacle position, resulting in an oscillatory trajectory.

Table~\ref{tab:real_results_rates} presents the quantitative results for the fixed-obstacle real-robot experiments, \added{demonstrating successful zero-shot transfer from simulation to the real robot.} Moreover, the EDF-RL-APF-FB policy demonstrated reactive avoidance in the interactive experiment with a dynamic object, without the need for fine-tuning. It is worth noting that given the collisions were considered for the full body of the robot, it is expected that the EE-only policies performed worse than the policies that considered points along the rest of the arm. Finally, as mentioned before, the EDF is constructed using data from the camera. Given that the field of view of the camera is facing the robot, there are many occlusions, in particular behind the obstacles. This missing information is the cause of some of the reported collisions with the arm.

\section{Conclusion}
% The conclusion will be finalised after completing the full evaluation and discussion.
We proposed a method for acquiring a versatile motion policy capable of adapting to unseen environments using RL. Specifically, we integrated an environment representation based on EDF into the PPO DRL algorithm. This method incorporates the EDF to provide real-time information on the distance to the nearest obstacle surface and the gradient direction for avoidance, based on the proximity to any point in the environment. This EDF information is used for observation and reward design in APF-guided configurations, making it a key feature of the proposed approach.
% This method extends the traditional APF approach by incorporating the EDF to provide real-time information on the distance to the nearest obstacle surface and the gradient direction for avoidance, based on the proximity to any point in the environment. This EDF information is used for both observation and reward design, making it a key feature of the proposed approach.
The experimental results show that using the distance and gradient information from the EDF enables the robot to exhibit stable avoidance \added{behaviour} in complex environments, achieving a high goal-reaching rate while reducing collisions. Additionally, the learned policy can be applied to environments with randomly changing obstacle configurations, resulting in a robust policy with high environmental adaptability.
Thus, the EDF-based environment representation is a promising approach for acquiring general-purpose motion policies through RL.

We have also identified several future developments. First, our real-world experiments used a single RGB-D camera, which limited obstacle reconstruction due to occlusions. As a next step, we will introduce multi-view sensing with multiple RGB-D cameras to obtain denser and more complete point clouds, enabling more reliable EDF construction and more rigorous real-world validation.

Second, we will extend the EDF-based representation beyond obstacle avoidance to grasping.
Existing grasping approaches often rely on specialised hardware like soft robots 
%(e.g., compliant hands or soft mechanisms)
% ~\cite{samadikhoshkho2019brief}
or data-driven grasp planners trained on large-scale datasets, % ~\cite{guo2016object},
whose performance can be sensitive to dataset scale and quality and can incur high data-collection costs. %~\cite{kappler2015leveraging}.

\section*{\added{ACKNOWLEDGMENT}}
\added{This research was mainly supported by JSPS KAKENHI Scientific Research (S) Grant Number 22H05002 and by the Industrial Transformation Training Centre (ITTC) for Collaborative Robotics in Advanced Manufacturing (also known as the Australian Cobotics Centre) funded by ARC (ProjectID: IC200100001).}

% \bibliographystyle{IEEEtran2etal}
% \bibliography{IROS2026_RL_EDF}

\added{
\bibliographystyle{IEEEtran}
\bibliography{IROS2026_RL_EDF}

@article{le2023accurate,
  title={Accurate gaussian-process-based distance fields with applications to echolocation and mapping},
  author={Le Gentil, Cedric and Ouabi, Othmane-Latif and Wu, Lan and Pradalier, Cedric and Vidal-Calleja, Teresa},
  journal={IEEE Robotics and Automation Letters},
  volume={9},
  number={2},
  pages={1365--1372},
  year={2023},
  publisher={IEEE}
}

@incollection{khatib1986potential,
  title={The potential field approach and operational space formulation in robot control},
  author={Khatib, Oussama},
  booktitle={Adaptive and Learning Systems: Theory and Applications},
  pages={367--377},
  year={1986},
  publisher={Springer}
}

@article{schneider2025robot,
  title={Robot movement planning for obstacle avoidance using reinforcement learning},
  author={Schneider, Linda-Sophie and Peng, Junyan and Maier, Andreas},
  journal={Scientific Reports},
  volume={15},
  number={1},
  pages={32506},
  year={2025},
  publisher={Nature Publishing Group UK London}
}

@book{sutton1998reinforcement,
  title={Reinforcement learning: An introduction},
  author={Sutton, Richard S and Barto, Andrew G and others},
  volume={1},
  number={1},
  year={1998},
  publisher={MIT press Cambridge}
}

@article{schulman2017proximal,
  title={Proximal policy optimization algorithms},
  author={Schulman, John and Wolski, Filip and Dhariwal, Prafulla and Radford, Alec and Klimov, Oleg},
  journal={arXiv preprint arXiv:1707.06347},
  year={2017}
}

@article{wu2021faithful,
  title={Faithful Euclidean distance field from log-Gaussian process implicit surfaces},
  author={Wu, Lan and Lee, Ki Myung Brian and Liu, Liyang and Vidal-Calleja, Teresa},
  journal={IEEE Robotics and Automation Letters},
  volume={6},
  number={2},
  pages={2461--2468},
  year={2021},
  publisher={IEEE}
}

@article{lavalle1998rapidly,
  title={Rapidly-exploring random trees: A new tool for path planning},
  author={LaValle, Steven},
  journal={Research Report 9811},
  year={1998},
  publisher={Department of Computer Science, Iowa State University}
}

@article{zhang2020optimization,
  title={Optimization-based collision avoidance},
  author={Zhang, Xiaojing and Liniger, Alexander and Borrelli, Francesco},
  journal={IEEE Transactions on Control Systems Technology},
  volume={29},
  number={3},
  pages={972--983},
  year={2020},
  publisher={IEEE}
}

@ARTICLE{Ali2024IDMP,
  author={Ali, Usama and Wu, Lan and Müller, Adrian and Sukkar, Fouad and Kaupp, Tobias and Vidal-Calleja, Teresa},
  journal={IEEE Robotics and Automation Letters}, 
  title={Interactive Distance Field Mapping and Planning to Enable Human-Robot Collaboration}, 
  year={2024},
  volume={9},
  number={12},
  pages={10850-10857},
  doi={10.1109/LRA.2024.3482128}}

@Article{app14114936,
AUTHOR = {Xu, Qiaoyu and Zhang, Tianle and Zhou, Kunpeng and Lin, Yansong and Ju, Wenhao},
TITLE = {Active Collision Avoidance for Robotic Arm Based on Artificial Potential Field and Deep Reinforcement Learning},
JOURNAL = {Applied Sciences},
VOLUME = {14},
YEAR = {2024},
NUMBER = {11},
ARTICLE-NUMBER = {4936},
URL = {https://www.mdpi.com/2076-3417/14/11/4936},
ISSN = {2076-3417},
DOI = {10.3390/app14114936}
}

@article{GOVONI2026105394,
title = {Real-time collision avoidance with robot distance fields in a task-priority framework},
journal = {Robotics and Autonomous Systems},
volume = {200},
pages = {105394},
year = {2026},
issn = {0921-8890},
doi = {https://doi.org/10.1016/j.robot.2026.105394},
url = {https://www.sciencedirect.com/science/article/pii/S0921889026000679},
author = {Andrea Govoni and Michela Cavuoto and Yiming Li and Sylvain Calinon and Gianluca Palli}
}

@article{article,
author = {Li, Haoxuan and Gong, Daoxiong and Yu, Jianjun},
year = {2021},
month = {06},
pages = {},
title = {An obstacles avoidance method for serial manipulator based on reinforcement learning and Artificial Potential Field},
volume = {5},
journal = {International Journal of Intelligent Robotics and Applications},
doi = {10.1007/s41315-021-00172-5}
}

@INPROCEEDINGS{8665083,
  author={Xu, Shan and Li, Gaofeng and Liu, Jingtai},
  booktitle={2018 IEEE International Conference on Robotics and Biomimetics (ROBIO)}, 
  title={Obstacle Avoidance for Manipulator with Arbitrary Arm Shape Using Signed Distance Function}, 
  year={2018},
  volume={},
  number={},
  pages={343-348},
  doi={10.1109/ROBIO.2018.8665083}}

@ARTICLE{Kavraki96,
  author = {L. Kavraki and P. Svestka and J.-C. Latombe and M. Overmars},
  title = {Probabilistic Roadmaps for Path Planning in High-Dimensional Configuration
	Spaces},
  journal = {IEEE Trans. on Robotics and Automation},
  year = {1996},
  volume = {12},
  number = {4},
  pages = {566-580}
}

@ARTICLE{Yip2020MPNet,
  author={Qureshi, Ahmed Hussain and Miao, Yinglong and Simeonov, Anthony and Yip, Michael C.},
  journal={IEEE Transactions on Robotics}, 
  title={Motion Planning Networks: Bridging the Gap Between Learning-Based and Classical Motion Planners}, 
  year={2021},
  volume={37},
  number={1},
  pages={48-66},
  doi={10.1109/TRO.2020.3006716}}

@INPROCEEDINGS{8202133,
  author={Tobin, Josh and Fong, Rachel and Ray, Alex and Schneider, Jonas and Zaremba, Wojciech and Abbeel, Pieter},
  booktitle={2017 IEEE/RSJ International Conference on Intelligent Robots and Systems (IROS)}, 
  title={Domain randomization for transferring deep neural networks from simulation to the real world}, 
  year={2017},
  volume={},
  number={},
  pages={23-30},
  doi={10.1109/IROS.2017.8202133}}
}

\end{document}